\documentclass[lettersize,journal]{IEEEtran}
\ifCLASSOPTIONcompsoc
  \usepackage[nocompress]{cite}
\else
  \usepackage{cite}
\fi
\ifCLASSINFOpdf
  \usepackage[pdftex]{graphicx}
\else
\fi
\usepackage{amsmath}
\usepackage{bbm}
\usepackage{float}
\usepackage{algorithmic}
\usepackage{array}
\ifCLASSOPTIONcompsoc
  \usepackage[caption=false,font=footnotesize,labelfont=sf,textfont=sf]{subfig}
\else
  \usepackage[caption=false,font=footnotesize]{subfig}
\fi
\usepackage{fixltx2e}
\usepackage{stfloats}

\usepackage{url}
\usepackage{comment}
\usepackage{amsmath,amsfonts}
\usepackage{array}
\usepackage{booktabs}
\usepackage{multirow}
\usepackage{hyperref}
\hypersetup{colorlinks,linkcolor={red},citecolor={cyan},urlcolor={magenta}} 
\usepackage{breakcites}
\usepackage[flushleft]{threeparttable}
\usepackage[misc,geometry]{ifsym}
\usepackage[table]{xcolor}
\usepackage{enumitem}
\usepackage{overpic}
\usepackage{bm}
\usepackage{float}
\usepackage{ragged2e}
\usepackage{xspace}
\usepackage{makecell} 
\usepackage{amssymb}
\usepackage{amsmath}
\usepackage{pifont}
\definecolor{mygray}{gray}{.92}
\definecolor{darkpastelgreen}{rgb}{0.01, 0.75, 0.24}
\definecolor{darkpastelred}{RGB}{230, 50, 0}
\definecolor{darkpink}{rgb}{0.91, 0.33, 0.5}
\definecolor{LightCyan}{rgb}{0.88,1,1}
\definecolor{Gray}{gray}{0.90}

\makeatletter
\DeclareRobustCommand\onedot{\futurelet\@let@token\@onedot}
\def\@onedot{\ifx\@let@token.\else.\null\fi\xspace}

\def\eg{\emph{e.g}\onedot} 
\def\ie{\emph{i.e}\onedot}

\def\wrt{w.r.t\onedot} 

\makeatother

\begin{document}

\title{Visual-geometric Collaborative Guidance for Affordance Learning}


\author{Hongchen Luo$^{1,2}$,
        Wei Zhai$^{1}$,
        Jiao Wang$^{2}$,
        Yang Cao$^{1}$ and Zheng-Jun Zha$^{1}$ \\ 

\IEEEcompsocitemizethanks{
\IEEEcompsocthanksitem $^{\textbf{1}}$ University of Science and Technology of China, Hefei, China
\IEEEcompsocthanksitem  $^{\textbf{2}}$ Northeastern University, Shenyang, China
\IEEEcompsocthanksitem Hongchen Luo (luohongchen@ise.neu.edu.cn), Wei Zhai (wzhai056@ustc.edu.cn), Jiao Wang (wangjiao@ise.neu.edu.cn), Yang Cao (forrest@ustc.edu.cn), Zheng-Jun Zha (zhazj@ustc.edu.cn)
\IEEEcompsocthanksitem A preliminary version of this work has appeared in 
CVPR 2023~\cite{luo2023leverage}.
}
}

\maketitle

\begin{abstract}
Perceiving potential ``action possibilities'' (\ie, affordance) regions of images and learning interactive functionalities of objects from human demonstration is a challenging task due to the diversity of human-object interactions. Prevailing affordance learning algorithms often adopt the label assignment paradigm and presume that there is a unique relationship between functional region and affordance label, yielding poor performance when adapting to unseen environments with large appearance variations. In this paper, we propose to leverage interactive affinity for affordance learning, \ie extracting interactive affinity from human-object interaction and transferring it to non-interactive objects. Interactive affinity, which represents the contacts between different parts of the human body and local regions of the target object, can provide inherent cues of interconnectivity between humans and objects, thereby reducing the ambiguity of the perceived action possibilities. To this end, we propose a visual-geometric collaborative guided affordance learning network that incorporates visual and geometric cues to excavate interactive affinity from human-object interactions jointly. Particularly, a semantic-pose heuristic perception (SHP) module is devised to exploit both semantic and geometric cues to guide the network to focus on interaction-relevant regions, alleviating the effects of combinatorial relational ambiguity. Meanwhile, A geometric-apparent alignment transfer module is introduced to jointly align local regions of apparent and structural similarity, eliminating the transport difficulties posed by intra-class correspondence ambiguity. Besides, a contact-driven affordance learning (CAL) dataset is constructed by collecting and labeling over 55,047 images. Experimental results demonstrate that our method outperforms the representative models regarding objective metrics and visual quality. Project: \href{https://github.com/lhc1224/VCR-Net}{github.com/lhc1224/VCR-Net}.
\end{abstract}

%}

\IEEEpeerreviewmaketitle

\section{Introduction}
\IEEEPARstart{T}{he} objective of affordance learning is to locate the ``action possibilities'' regions of an object \cite{gibson1977theory}, which is crucial in the embodied intelligence field. For an intelligent agent in an interacting environment, it is vital to perceive not only the object semantics but also how to interact with various objects' local regions. Perceiving and reasoning about the object's interactable regions is a critical capability for embodied intelligent systems to interact with the environment actively, distinct from passive perception systems \cite{nagarajan2019grounded,nagarajan2020learning}. Moreover, affordance learning has a wide range of applications in fields such as action recognition \cite{qi2017predicting}, scene understanding \cite{chuang2018learning}, human-robot interaction \cite{yamanobe2017brief}, autonomous driving \cite{chen2015deepdriving} and VR/AR \cite{sun2021impression}.

\begin{figure}[t]
	\centering
		\begin{overpic}[width=0.97\linewidth]{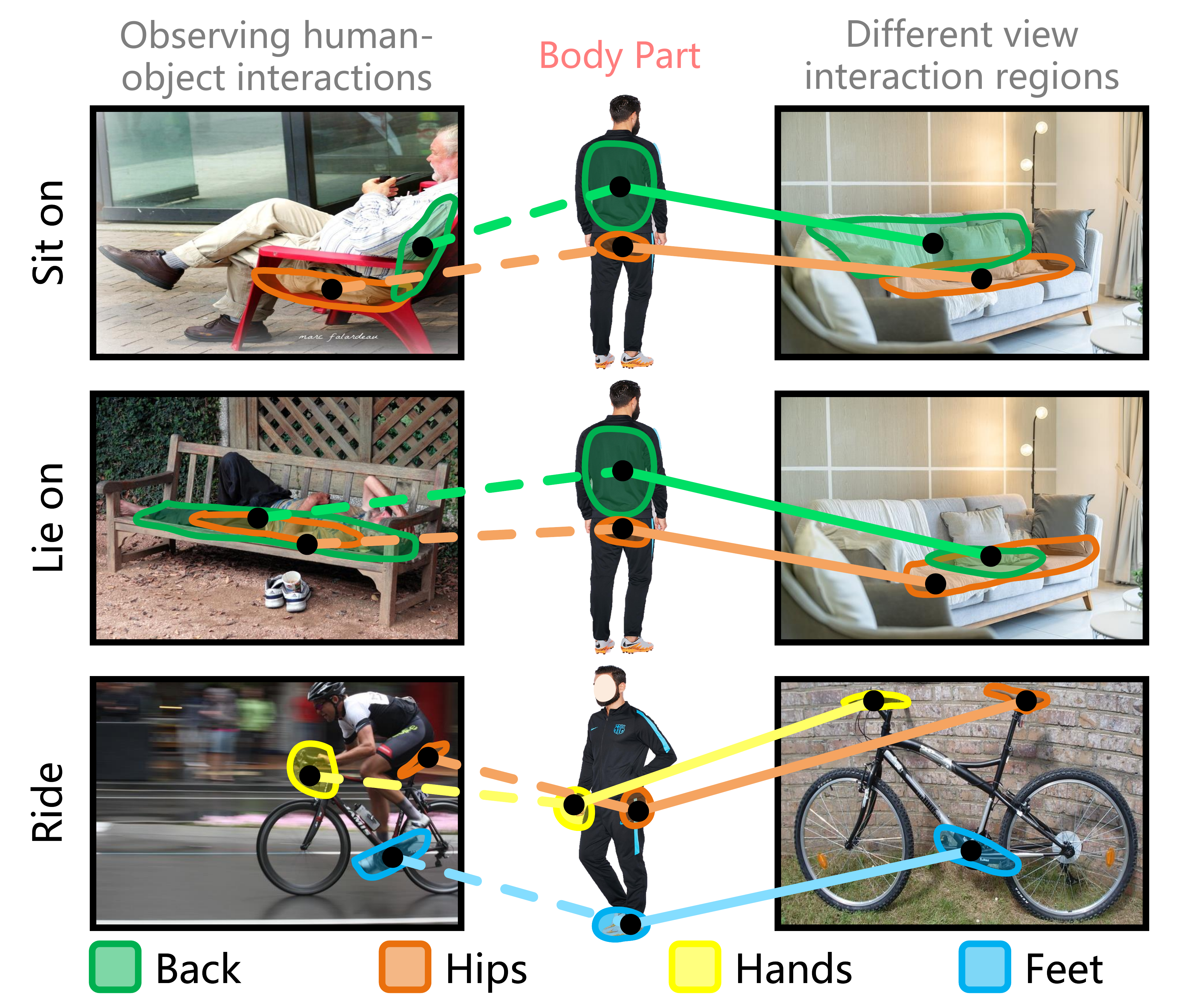}
	\end{overpic}
	\vspace{-10pt}
	\caption{\textbf{Interactive affinity. }\textbf{(a)} Interaction affinity refers to the contact between different parts of the human body and the local regions of a target object. \textbf{(b)} The interactive affinity provides rich cues to guide the model to acquire invariant features of the object's local regions interacting with the body part, thus counteracting the multiple possibilities caused by diverse interactions. }
	\label{fig1}
\end{figure}

\par Affordance is a dynamic property closely related to humans and the environment  \cite{hassanin2021visual}. Previous works \cite{do2018affordancenet,myers2015affordance,sawatzky2017adaptive,nguyen2016detecting} focus on establishing mapping relationships between appearances and labels for affordance learning. However, they neglect the multiple possibilities of affordance brought about by changes in the environment and actors, leading to an incorrect perception. Recent studies \cite{nagarajan2020learning,schiavi2022learning,ning2024where2explore} utilize reinforcement learning to allow intelligent agents to perceive the environment through numerous interactions in simulated/actual scenarios. Such approaches are mainly limited by their high cost and struggle to generalize to unseen scenarios \cite{wang2022adaafford}. To this end, researchers consider learning from human demonstration in an action-free manner \cite{nagarajan2019grounded,luo2021one,luo2022learning,fang2018demo2vec,yang2024learning}. Nonetheless, they only roughly segment the whole object/interaction regions in a general way, which is still challenging to understand how the object is used. The multiple possibilities due to different local regions interacting with humans in various ways are not fully resolved.  In this paper, we propose to leverage interactive affinity for affordance learning,  \ie extracting interactive affinity from human-object interaction and transferring it to non-interactive objects. The interactive affinity (as shown Fig. \ref{fig1} (a)) denotes the contacts between different human body parts and objects' local regions, which can provide inherent cues of interconnectivity between humans and objects, thereby reducing the ambiguity of the perceived action possibilities (as shown in Fig. \ref{fig1} (b)).

\begin{figure}[t]
	\centering
		\begin{overpic}[width=0.955\linewidth]{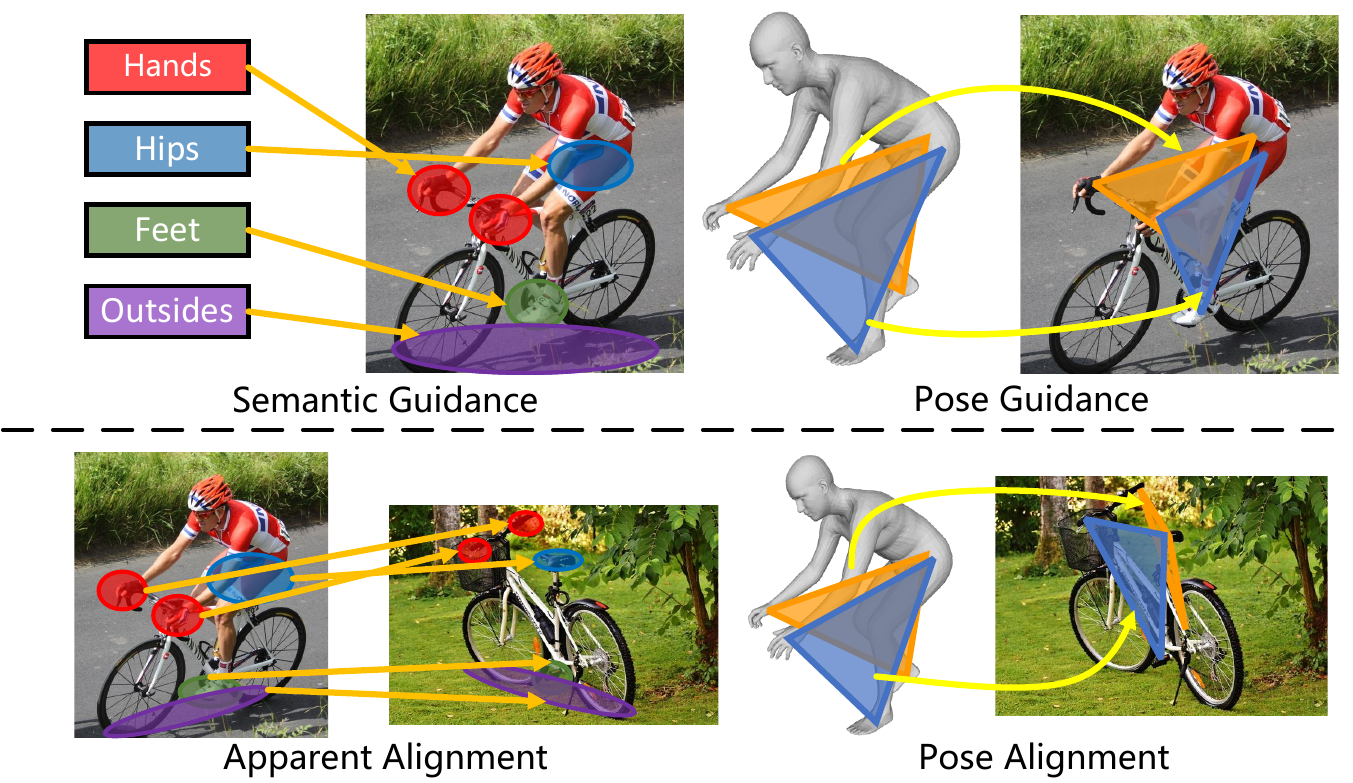}
   \put(0.3,28){\textbf{(a)}}
  \put(0.3,22){\textbf{(b)}}
	\end{overpic}
 \vspace{-10pt}
	\caption{\textbf{Motivation.} (a) We consider both the semantic and structural cues to extract the interactive affinity from the interaction images. (b) We exploit the implicit structural cues of body pose and apparent similarity to jointly perform the interactive affinity transfer.}
	\label{motivation}
\end{figure}

However, multiple ambiguities in the interaction process render it challenging for the model to perceive and robustly generalize the interactive affinity representation. This is mainly reflected in the ambiguity of combinatorial relations and intra-class correspondence. The combinatorial relationship ambiguity means that due to the diversity of human-object interactions, the combination of interactions between the body and the object's local regions is complex and various, resulting in the model's difficulty in perceiving the contact regions corresponding to distinct interactions and accurately mining the interactive affinity representations. To address this problem, we consider leveraging semantic and geometric cues to guide the model in mining interactive affinity (Fig. \ref{motivation} (a)). We leverage the textual semantics to explicitly align visual features to guide the model to focus on interaction-related contact regions while exploiting the geometric relationships between various body joints to mine potential structural cues in the interactable regions of an object, enabling the model to focus on complex interactive combinatorial relationships adequately. The intra-class correspondence ambiguity refers to the fact that since the same affordance covers multiple classes of objects, there are significant variations in the appearance,  views, and scales, which makes the representations of the interactable local regions corresponding to the objects and the relative spatial relationships in the interactive and non-interactive images more inconsistent, resulting in the possible occurrence of negative transfer. To this end, we consider both apparent and geometric alignment to achieve accurate transfer (Fig. \ref{motivation} (b)). We use dense matching to align visual similarity features between local interactable regions while utilizing the geometric structure of the human pose as a bridge to align the similar geometric structure of interactable regions, eliminating the effects of intra-class corresponding ambiguity by exploiting the complementary relationships.

In this paper, we propose a \textbf{V}isual-geometric \textbf{C}ollaborative guided affo\textbf{R}dance learning \textbf{Net}work (\textbf{VCR-Net}) for extracting interactive affinity representations in interactive images and transferring them to non-interactive images to achieve understanding and generalization of various complex interactions. Firstly, a \textbf{S}emantic-pose \textbf{H}euristic \textbf{P}erception (\textbf{SHP}) module is designed to guide the model to focus on interaction-relevant local contact regions, obtaining interactive affinity representations from diverse interactions. Subsequently,  a \textbf{G}eometric-apparent \textbf{A}lignment \textbf{T}ransfer (\textbf{GAT}) module is introduced to transfer the interactive affinity to the non-interactive image. Specifically, the SHP module utilizes a cross-transformer to inject the semantic cues from different body parts into the interactive image's features. Meanwhile, it employs the deep equilibrium model \cite{bai2019deep} to adaptively adjust the geometrical correspondences between the image and mesh features, enabling the model to adapt to various complex postures. On the other hand, the GAT module adaptively adjusts the association between mesh and non-interactive image features using the deep equilibrium model during the interactive affinity representation transfer process. and then it considers the contact region features extracted from the interactive image and the non-interactive features to calculate the feature similarity between pixels to activate the corresponding interactable local regions.  

\par Although the numerous related datasets \cite{zhai2022one,luo2022learning,luo2023grounded,chuang2018learning,wang2017binge,yang2024learning} that emerged during the development of affordance learning, there is still a lack of relevant datasets suited for leveraging interactive affinity. To carry out a thorough research, this paper constructs an  \textbf{C}ontact-driven \textbf{A}ffordance \textbf{L}earning (\textbf{CAL}) dataset, consisting of $55,047$ images from $35$ affordance and $61$ object categories. We conduct contrastive studies on the CAL dataset against eight representative models in several related fields. Experimental results validate the effectiveness of our method in solving the multiple possibilities of affordance. In summary, our primary contributions are:
\par\textbf{1)} This paper considers leveraging interactive affinity for affordance learning and establishing a CAL benchmark to facilitate the study of obtaining interactive affinity to counteract the multiple possibilities of affordance.
\par\textbf{2)} A novel visual-geometric collaborative guided affordance learning network is proposed that utilizes visual and geometric cues to achieve more accurate interactive affinity representations extracted from human-object interactions, eliminating the effects of multiple ambiguities.  
\par\textbf{3)}   Experiments on the CAL dataset demonstrate that our VCR-Net outperforms state-of-the-art methods and can serve as a strong baseline for affordance learning research.

This paper builds upon our conference version \cite{luo2023leverage}, which has been extended in three distinct aspects. \textbf{Firstly}, we provide an in-depth analysis of the interactive affinity extraction and the transfer process. \textbf{Secondly}, we introduce a semantic-pose heuristic perception module that leverages semantic and geometric structures jointly to guide the model to extract the interactive affinity representation more accurately. \textbf{Thirdly}, we introduce a geometric-apparent alignment transfer module to eliminate inconsistencies during transfer by aligning structural and apparent similarities. \textbf{Fourthly}, we expand the dataset nearly ten times in several aspects and conduct more experiments to analyze the model's performance comprehensively.

\par The remainder of this paper is organized as follows: Sec. \ref{related works} provides a brief review of existing related studies. Sec. \ref{section_method} describes the pipeline of the proposed model and its details. In Sec. \ref{section_dataset}, we introduce the collection, annotation process, and statistical analysis of the CAL dataset. Sec. \ref{sec:experiments} describes the experimental setting and provides comprehensive results and analysis. In Sec. \ref{sec:Conclusion and Discussion}, we present the conclusions, limitations, and future directions of this work.

\section{Related Work}
\label{related works}
\subsection{Affordance Learning}
In recent years, the rapid growth of robotics, autonomous driving, and embodied intelligence has thrust affordance learning into prominence \cite{hassanin2021visual}.  Early works \cite{do2018affordancenet,myers2015affordance,nguyen2016detecting,deng20213d,sawatzky2017weakly} mainly adopt the label assignment paradigm, positing an intrinsic correlation between functional regions and their corresponding affordance labels. However, it encounters challenges in addressing the multiplicity of affordance interpretations, which arise from variations in environmental contexts and operator behavior. Recent studies have integrated reinforcement learning, leveraging tailored reward functions to enable agents to dynamically interact with their environment and refine their affordance perception skills \cite{nagarajan2020learning,liao2022rediscovering,schiavi2022learning}. This series of works can mitigate the problem of generalization to diverse environments. However, the actions and states are in a high-dimensional and complex space, which is difficult to optimize using autonomous exploration and is restricted by the scenarios of the simulation platform, which prevents it from being applied in various complex scenarios and tasks. Some studies \cite{geng2023gapartnet,geng2023sage,li2024manipllm} achieve generalization of articulated object manipulation by decoupling the object part while aligning the relationship between semantics and interactable parts. While some other works consider learning the object's affordance from the human demonstration in an action-free manner \cite{fang2018demo2vec,luo2021one,luo2022learning,zhai2022one,nagarajan2019grounded,Li_2023_CVPR,Yang_2023_ICCV,Chen_2023_CVPR}, extracting interactions from the images/videos and transferring the human action intentions implied within them to the new unseen object, thus achieving perception and generalization. Then,  Yang et al. \cite{yang2024learning} proposed a multi-task learning framework named GAAF-Dex that learns grasping knowledge from the exocentric view and transfers it to the egocentric view, further advancing interactive skill learning. However, they only detect/segment the object as a whole or the interactable regions in a general way. They do not perceive how the object's local regions are used and have yet to resolve the multiple possibilities issue fully. In contrast, this paper considers using the inherent cues of interconnectivity between humans and objects to reduce the ambiguity of the perceived action possibilities.

\subsection{Interaction Relationship Learning}
Since studying the human-object interaction relationship is crucial for improving several areas, such as the autonomy of intelligent agents and the interaction of environments, this has attracted a wide range of attention from researchers. Some work \cite{Gkioxari_2018_CVPR,chao2018learning,Kim_2021_CVPR} mainly considered detecting human-object interaction pairs from images. However, the diversity of interactions and the variability in actions for different objects complicate the mapping between interaction labels and visual cues, posing challenges for accurate perception and generalization. Some efforts consider contact relationships between human beings and objects as cues to achieve a more accurate understanding of interactions. Shimada et al. \cite{shimada2022hulc} use body-scene contacts to guide 3D human capture. Bhatnagar et al. \cite{bhatnagar2022behave} propose to jointly track humans, objects, and contacts along with collecting a large-scale BEHAVE dataset containing human models, objects, and contact annotations. Mao et al. \cite{mao2022contact} introduce distance-based contact maps as an explicit constraint for human motion forecasting. Yang et al. \cite{yang2021cpf} propose to model hand-object interaction by explicitly representing the contact using the Contact Potential Field (CPF). Similar to the above work, we utilize the interactive affinity to mine the interconnectivity between humans and objects, helping reduce the ambiguity in affordance learning.

\section{Method}
\label{section_method}
\begin{figure*}[t]
	\centering
		\begin{overpic}[width=0.97\linewidth]{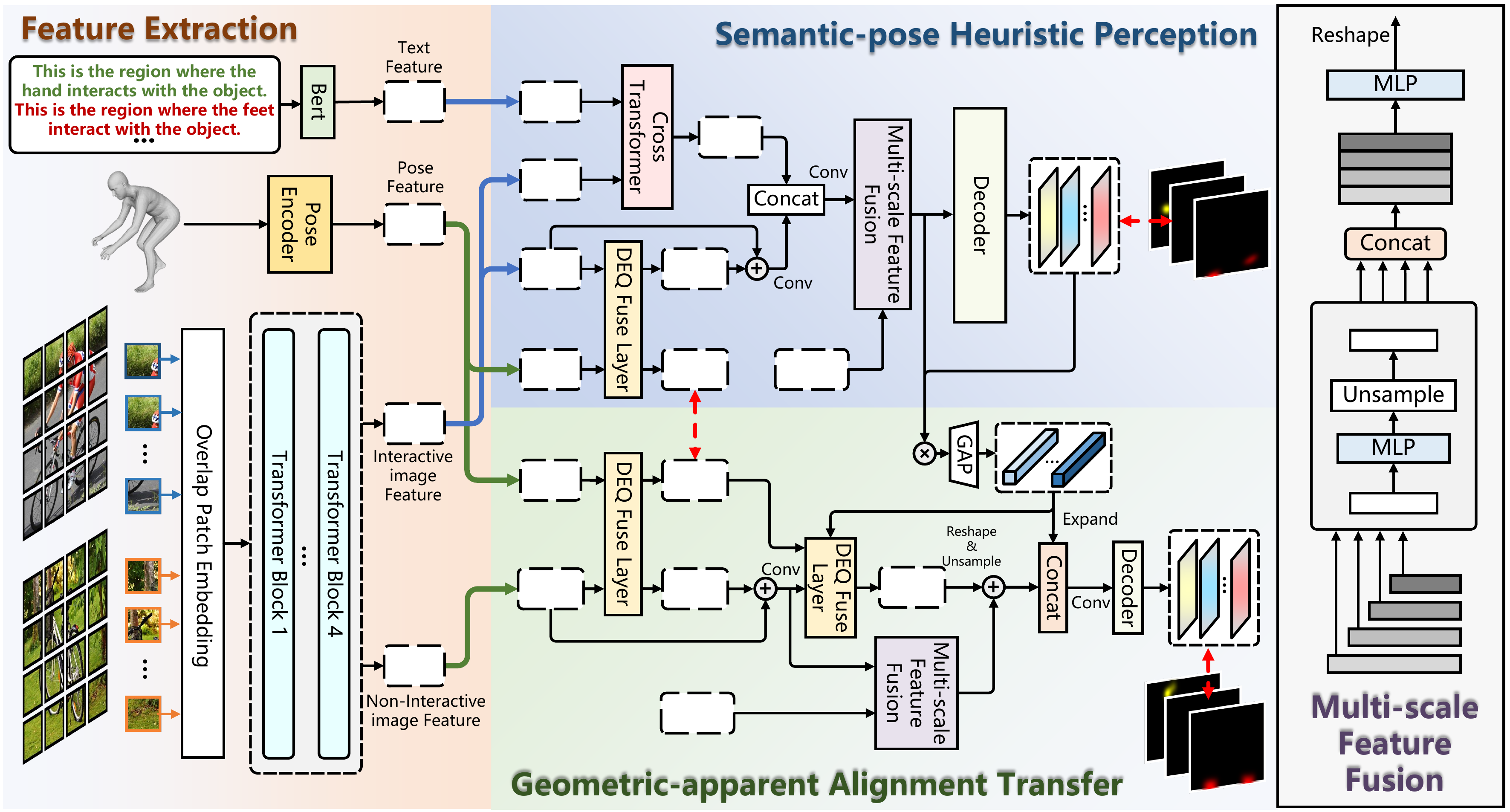}
  \scriptsize
     \put(26.2,46.38){$\bm{X}^T$}
     \put(26.2,38.3){$\bm{X}^P$}
     \put(26.3,25.1){{$\mathbb{X}^{in}$}}
     \put(25.9,9.1){{$\mathbb{X}^{non}$}}
      \put(35.3,46.48){$\bm{X}^T$}
      \put(35.1,41.35){{$\bm{X}^{in}_4$}}
      \put(35.1,35.55){{$\bm{X}^{in}_4$}}
       \put(35.3,28.65){$\bm{X}^{P}$}
        \put(35.28,21.4){$\bm{X}^{P}$}
        \put(34.5,14.35){$\bm{X}^{non}_4$}
        \put(46.92, 44.1){$\bm{\hat{X}}^{in}$}
        \put(44.72,35.35){$\bm{Z}^{in}$}
        \put(45.,28.75){$\bm{Z}^P$}
        \put(47.1,25.2){$\bm{\mathcal{L}}_{align}$}
         \put(45.2,21.4){$\bm{\bar{Z}}^{P}$}
         \put(44.15,14.2){$\bm{Z}^{non}$}
         \put(44.03, 6.13){$\bm{X}^{non}_{2,3,4}$}
          \put(51.55,28.93){$\bm{X}^{in}_{2,3,4}$}
          \put(58.61,14.34){$\bm{\hat{Z}}^{non}$}
          \put(74.4,35.5){$\bm{\mathcal{L}}_{in}$}
          \put(76.,9.78){$\bm{\mathcal{L}}_{non}$}
	\end{overpic}
 \vspace{-9pt}
\caption{\textbf{Overview of the proposed VCR-Net.} Our approach mainly consists of three parts: feature extraction, semantic-pose heuristic perception module (Sec. \ref{sec:IAE}) and geometric-apparent alignment transfer module (Sec. \ref{sec:IAT}).}
	\label{pipeline}
\end{figure*}

Given a human-object interaction image $\bm{I}^{in}$ with a corresponding human pose $\bm{P}$ and a non-interactive image $\bm{I}^{non}$, we aim to extract the affordance affinity representation between the human body part and the object local region from $I^{in}$ and transfer it to $I^{non}$ to predict the corresponding interactable region. The \textbf{VCR-Net} is shown in Fig. \ref{pipeline}. It first extracts features through a transformer \cite{xie2021segformer} backbone to obtain $\mathbb{X}^{in}=\{\bm{X}^{in}_i,i\in[1,4]\}$ and $\mathbb{X}^{non}=\{\bm{X}^{non}_i,i\in[1,4]\}$, respectively ($i$ indexes the block of the backbone). For human poses, we use a pose encoder consisting of several layers of transformers \cite{dosovitskiy2020image} to obtain the feature representation $\bm{X}^P$. For different body part descriptions, the text features are extracted using Bert \cite{turc2019}: $\bm{X}^T=\text{Bert}(\bm{T})$. Subsequently, an semantic-pose heuristic perception (\textbf{SHP}) module collectively guides the network through semantics and pose to obtain affinity representations from human-object interaction images (Sec. \ref{sec:IAE}). Finally, the geometric-apparent alignment transfer (\textbf{GAT}) module guides the network to transfer the interactive affinity representation to the non-interactive image through the geometric prior of the pose and the local region's apparent similarity (Sec. \ref{sec:IAT}).

\subsection{Semantic-pose Heuristic Perception}
\label{sec:IAE}
Due to the complexity and variety of human-object interaction relationships, there exists combinatorial ambiguity between human-object interaction contacts, rendering it difficult to perceive the interactive affinity accurately. However, the collaborative relationships between body parts are similar to human-object interactions. To this end, we consider using the rotational offset relationship of different joint locations in the 3D mesh \cite{cai2024smpler} to direct the model's focus on the corresponding interactable regions. Meanwhile, to strengthen the module's ability to perceive different body parts, we introduce semantic guidance, which enhances the network's perception of the affinity representation of different parts with objects by aligning the association between semantic descriptions of different parts and visual representations of different parts in the feature space during the training process.  As shown in Fig. \ref{pipeline}, firstly, we exploit the cross-transformer \cite{dosovitskiy2020image} to align the connection between semantic and visual features of the interactive images which enhance the model's perception of the different body part contact regions. The cross transformer ($\bm{O} =\text{CT}(\bm{X}_1,\bm{X}_2)$) is computed as:
\begin{equation}
\small
    \bm{Y}=\text{MCA}(\text{LN}(\bm{X}_1),\text{LN}(\bm{X}_2))+\bm{X}_1,
\bm{O}=\text{MLP}(\text{LN}(\bm{Y}))+\bm{Y},
\end{equation}
where $\bm{X}_1$ represents query, $\bm{X}_2$ represents key and value, $\text{MCA}()$ denotes the dot-production attention \cite{vaswani2017attention}. In the SHP module, we concatenate the features of text and interactive images as key and value, $\text{I}^{in}$ as query fed into a one-layer cross transformer to guide the network to establish the connection between different body parts and visual features: 
\begin{equation}
    \small
    \bm{\hat{X}}^{in}=\text{CT}(\bm{X}^{in},[\bm{X}^{in},\bm{X}^T]). 
\end{equation}
where $[\cdot, \cdot]$ represents the concatenate operation. On the other hand, human joints can provide rich geometric cues for the model to mine associations between the interactable local regions of an object. However, due to the large modal differences between pose features and image features and the large differences in the interactive images in various views, it is arduous to learn the links between the two modalities with a fixed network layer. To this end, we adaptively adjust the correspondence between the joints and the local regions of the objects in the interactive images by introducing the Deep Equilibrium Model \cite{bai2019deep} to dynamically sense effective information and remove redundancy. In general, the DEQ model forward process can be written:
\begin{equation}
    \small
\bm{Z}^{[i+1]}=f_\theta(\bm{Z}^{[i]};\bm{X}), \quad i=0,..,L-1.
\end{equation}
DEQ denotes an infinitely deep network with only one layer of $f_\theta$ that converges to the equilibrium state and can be implicitly backpropagated in one calculation. The results are independent of the chosen root-finding algorithm or the network structure of $f_\theta$, which provides more flexibility in the design of $f_\theta$. In the SHP module, for a given input $\bm{X}=\{\bm{X}_i\}$, we define the following form of $f$ with $i\in[1,2]$ for example, $\bm{Q}$ is calculated as follows:
\begin{equation}
    \small
    \bm{Q}=[\bm{W}^{Q}_1\bm{X}_1,\bm{W}^{Q}_2\bm{X}_2]+[\bm{W}^{Q'}_1\bm{Z}_1,\bm{W}_2^{Q'}\bm{Z}_2],
\end{equation}
$\bm{K}$ and $\bm{V}$ are calculated similarly to $\bm{Q}$ and $f_\theta$ is expressed as follows:
\begin{equation}
    \small
    f_\theta(\bm{Z};\bm{X})=\text{FFN}(\bm{Z}+\text{Attention}(\bm{Q},\bm{K},\bm{V}))+ \bm{Z}).
\end{equation}
Specifically, in order to utilize pose to guide interactive and non-interactive images to establish connections between interactable regions, we employ the following format:
\begin{equation}
    \small
    \bm{Z}^{in*}=\text{DEQFuse}([\bm{X}^{in}_4,\bm{X}^P]).
\end{equation}
where $\bm{Z}^{in*}$ can be split into $\bm{Z}^P$ and $\bm{Z}^{in}$. We sum $\bm{Z}^{in}$ with the original $\bm{X}^{in}_4$ features to obtain an interactable region feature representation that fuses the pose cues. Then, it is concatenated with $\bm{X}^{in}_4$ and fed into a layer of convolution to obtain an affinity representation $\bm{X}^{in}_{sp}$ that fuses the pose geometry information and semantic cues:
\begin{equation}
    \small
\tilde{\bm{X}}^{in}=\text{Conv}(\bm{Z}^{in}+\bm{X}^{in}_4), \quad
    \bm{X}_{sp}^{in}=\text{Conv}([\tilde{\bm{X}}^{in},\hat{\bm{X}}^{in}]).
\end{equation}
After obtaining the active affinity representation, we send it to a multi-scale feature fusion layer to fuse with the shallow features extracted from the backbone to obtain the higher resolution features \cite{xie2021segformer}:
\begin{equation}
\small
\bm{F}^{in}_i=\text{Upsample}(\text{MLP}(\bm{X}_{sp}^{in})), \quad i=4, \label{eq14}
\end{equation}
\begin{equation}
\small
    \bm{F}^{in}_i=\text{Upsample}(\text{MLP}(\bm{X}^{in}_i)), \quad i\in [1,3],  \label{eq15}
\end{equation}
\begin{equation}
\small
    \bm{\hat{F}}^{in}=\text{MLP}([\bm{F}^{in}_1, \bm{F}^{in}_2, \bm{F}^{in}_3, \bm{F}^{in}_4]). \label{eq16}
\end{equation}
After fusing multi-scale features, $\bm{\hat{F}}^{in}$ is reshaped to 2D feature map and sent to a decoder $\text{Decoder}(\cdot)$ to obtain the contact prediction: $\bm{D}^{in}= \text{Decoder}(\bm{\hat{F}}^{in})$. $\bm{D}^{in}$ contains $N$ (the number of body parts) channels, which represents the regions of body parts interact with the object. 

\begin{figure}[t]
	\centering
		\begin{overpic}[width=0.92\linewidth]{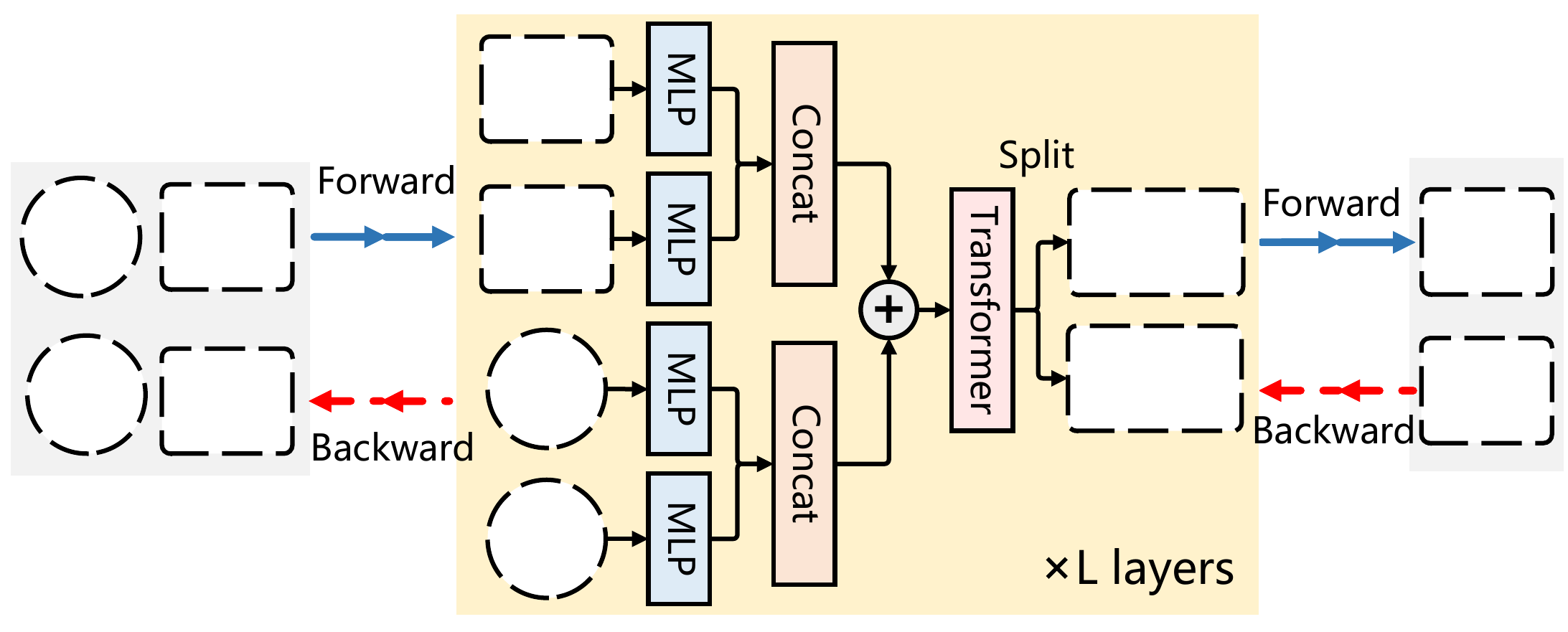}
  \scriptsize
  \put(3.,24.25){$\bm{X}_1$}
  \put(11.8,23.8){$\bm{Z}_1^{[0]}$}
  \put(3.,14.1){$\bm{X}_2$}
  \put(11.57,13.6){$\bm{Z}_2^{[0]}$}
  \put(32.8,14.5){$\bm{X}_1$}
  \put(32.1,33.3){$\bm{Z}_1^{[i]}$}
  \put(32.8,5.){$\bm{X}_2$}
  \put(32.1,23.9){$\bm{Z}_2^{[i]}$}
  \put(69.3,23.5){$\bm{Z}_1^{[i+1]}$}
  \put(69.3,14.65){$\bm{Z}_2^{[i+1]}$}
  \put(91.8,14.1){$\bm{Z}_2^{[L]}$}
   \put(91.8,23.4){$\bm{Z}_1^{[L]}$}
	\end{overpic}
 \vspace{-8pt}
	\caption{\textbf{DEQ fuse layer.} The DEQ fuse layer consists of a transformation $f_\theta$ that is driven to equilibrium between different input features.}
	\label{Deq}
\end{figure}
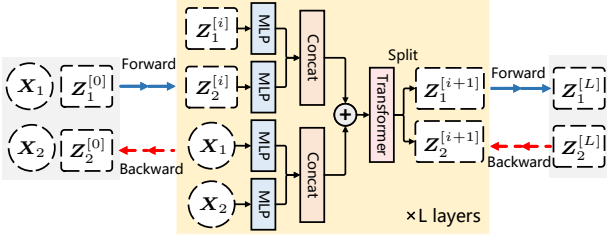

\subsection{Geometric-apparent Alignment Transfer}
\label{sec:IAT}
Due to multiple classes of objects in the same affordance category, apparent, viewpoint, and scale variations create intra-class correspondence ambiguities, resulting in inaccurate transfer of the interactive affinity. To this end, we align structural and apparent aspects to achieve a more accurate and robust transfer. As shown in Fig.\ref{pipeline}, for geometric correspondences, we first use the pose as a bridge to activate regions in the non-interactive image that can match the geometric structures presented by the different joints of the body. Similar to the operations in the SHP module where pose guides the interactable region in the interactive image, we use the DEQ fusion layer to establish the link between the non-interactive image features and the pose features:
$\bm{Z}^{non*}=\text{DEQFuse}([\bm{X}^{non}_4,\bm{X}^P])$.
where $\bm{Z}^{non*}$ can be split into $\bar{\bm{Z}}^P$ and $\bm{Z}^{non}$: $\bm{\tilde{X}}^{non}=\text{Conv}(\bm{Z}^{non}+\bm{X}^{non}_4)$. To bring the pose-guided interactions with the corresponding geometrical structures in the non-interactive images as close as possible, we introduce a pose alignment loss $\bm{\mathcal{L}}_{align}$ that is used to narrow down the features of the poses in the two DEQ fusion layers:
\begin{equation}
    \small
\bm{\mathcal{L}}_{align}=KLD(\bar{\bm{Z}}^P,\bm{Z}^P).
\end{equation}
For the apparent matching relationship, we first input the pose, $\bm{\tilde{X}}^{non}$, and $\bm{G}^{in}$ ($\bm{G}^{in}=[\bm{G}^{in}_1,...,\bm{G}^{in}_N]$, where $\bm{G}^{in}_j=\text{MASK}_i \otimes \hat{\bm{F}}^{in}$) together into the DEQ fusion layer, which utilises the pose and the corresponding mask region in $\bm{Z}^{non}$ to orientate the network to focus on the local features in the contact region that are related to the pose:
\begin{equation}
    \small
\bm{\hat{Z}}^{non*}=\text{DEQFuse}([\bm{\tilde{X}}^{non}, \bm{G}^{in}, \bar{\bm{Z}}^P].
\end{equation}
Non-interactive image features $\bm{\hat{Z}}^{non}$ can be obtained by $\bm{\hat{Z}}^{non*}$ split. Meanwhile, $\bm{\tilde{X}}^{non}$ take the feature representation of the non-interactive branch $\bm{Z}^{non}$  and compute the high-resolution feature output $\hat{\bm{F}}^{non}$ by following Eq. \ref{eq14} $\sim$ Eq. \ref{eq16}.
Subsequently, the pooled $\bm{G}_j^{in}$ is concatenated with $\bm{\hat{F}}^{non}$ and $\bm{R}^{non}$ and fed into the convolutional layer to activate the regions where the features are apparently similar and go through the decoder to obtain the final output:
\begin{equation}
    \small
   \bm{F}^{fuse}=\text{Conv}([\bm{\hat{F}}^{non}+\text{Upsample}(\bm{\hat{Z}}^{non}),\text{Expand}(\bm{G}^{in})]), 
\end{equation}
\begin{equation}
    \small
    \bm{D}^{non}=\text{Decoder}(\bm{F}^{fuse}).
\end{equation}

\begin{table}[!t]
\caption{\textbf{The dimensions, domains of definition, and meanings of the symbols used in the proposed approach.}
}
\vspace{-5pt}
\centering
 %\small
% \footnotesize
\scriptsize
 \renewcommand{\arraystretch}{1.}
  \renewcommand{\tabcolsep}{0.7pt}
  \begin{tabular}{c||c|c}
\hline
\Xhline{2.\arrayrulewidth}
     &  \textbf{Dimensions} &  \textbf{Meanings}  \\
\hline
\Xhline{2.\arrayrulewidth}
   $\bm{I}^{in}$/$\bm{I}^{non}$   & $3 \times 224 \times 224$   & Input image  \\
   $\bm{P}$ &  $53 \times 3$  & Human pose \\
   $\bm{X}_i^{in}$/$\bm{X}_i^{non}$ & $c_i \times h_i \times w_i$ & Image feature \\
   $\bm{X}^P$ & $c \times 53$ & Pose feature \\
   $\bm{Z}^P$ & $c \times 53$ & Pose features after DEQ fusion layer   \\
   $\bm{Z}^{in}$ & $c \times h_4w_4$ & Interactive features after DEQ fusion layer \\
   $\hat{\bm{F}}^{in}$ & $c \times h_1 \times w_1$ & Interactive features after multiscale fusion \\
   $\bm{Z}^{non}$ & $c \times h_4w_4$ & Non-interactive features after DEQ fusion layer \\
   $\bar{\bm{Z}}^{P}$ & $c \times 53$ & Pose features after DEQ fusion layer \\
   $\bm{G}^{in}$ & $c \times h_1 \times w_1$ & Interactive contact region features \\
   $\bm{\hat{Z}}^{non}$ & $c \times h_4w_4$ & Non-interactive features after DEQ fusion layer \\
   $\bm{\hat{F}}^{non}$ & $c \times h_1 \times w_1$ & Non-interactive features after multiscale fusion \\
   $\bm{D}^{in}$/$\bm{D}^{non}$ & $N_{cls} \times h_1 \times w_1$  & Contact region prediction \\

\hline
\Xhline{2.\arrayrulewidth}
    \end{tabular}
    \label{symbols}
\end{table}
During training, the total loss is:
$\bm{\mathcal{L}}_{total}=\lambda_1 \bm{\mathcal{L}}_{in}+\lambda_2\bm{\mathcal{L}}_{non}+\lambda_3\bm{\mathcal{L}}_{align}$,
where $\bm{\mathcal{L}}_{in}$ and $\bm{\mathcal{L}}_{non}$ are the binary cross-entropy loss for the two branches, respectively. $\lambda_1$, $\lambda_2$ and $\lambda_3$ are the loss weight parameters and are both set to $1$.
\section{Dataset}
\label{section_dataset}
This section introduces the collection, annotation, and attribute analysis of contact-driven affordance learning (CAL) datasets. In particular, section \ref{dataset collection} describes the sources, selection criteria, and filtering process. Section \ref{dataset ann} illustrates the annotation rules and processing. Section \ref{statistic} provides the properties of the CAL dataset and the results of the statistical analyses.

\begin{figure*}[t]
	\centering
		\begin{overpic}[width=0.97\linewidth]{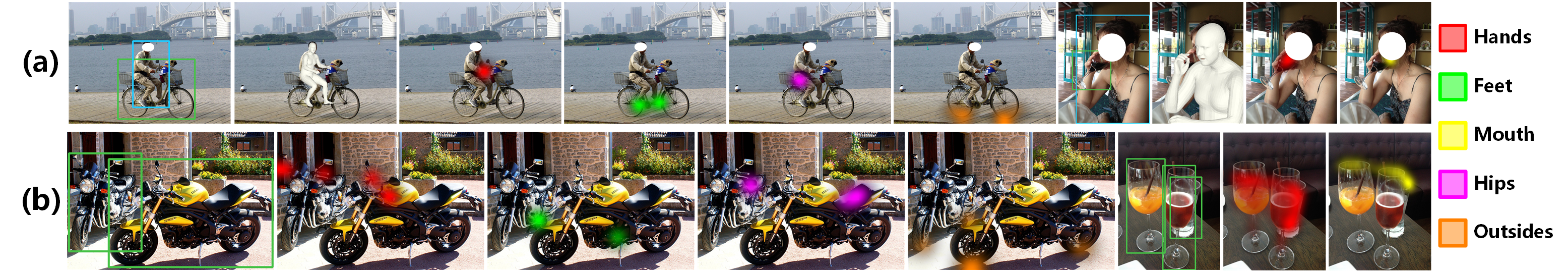}
	\end{overpic}
 \vspace{-8pt}
	\caption{\textbf{Dataset image examples.} Some examples of images and annotations from the CAL dataset.}
 \label{dataset_example}
\end{figure*}

\begin{figure*}[t]
	\centering
		\begin{overpic}[width=0.96\linewidth]{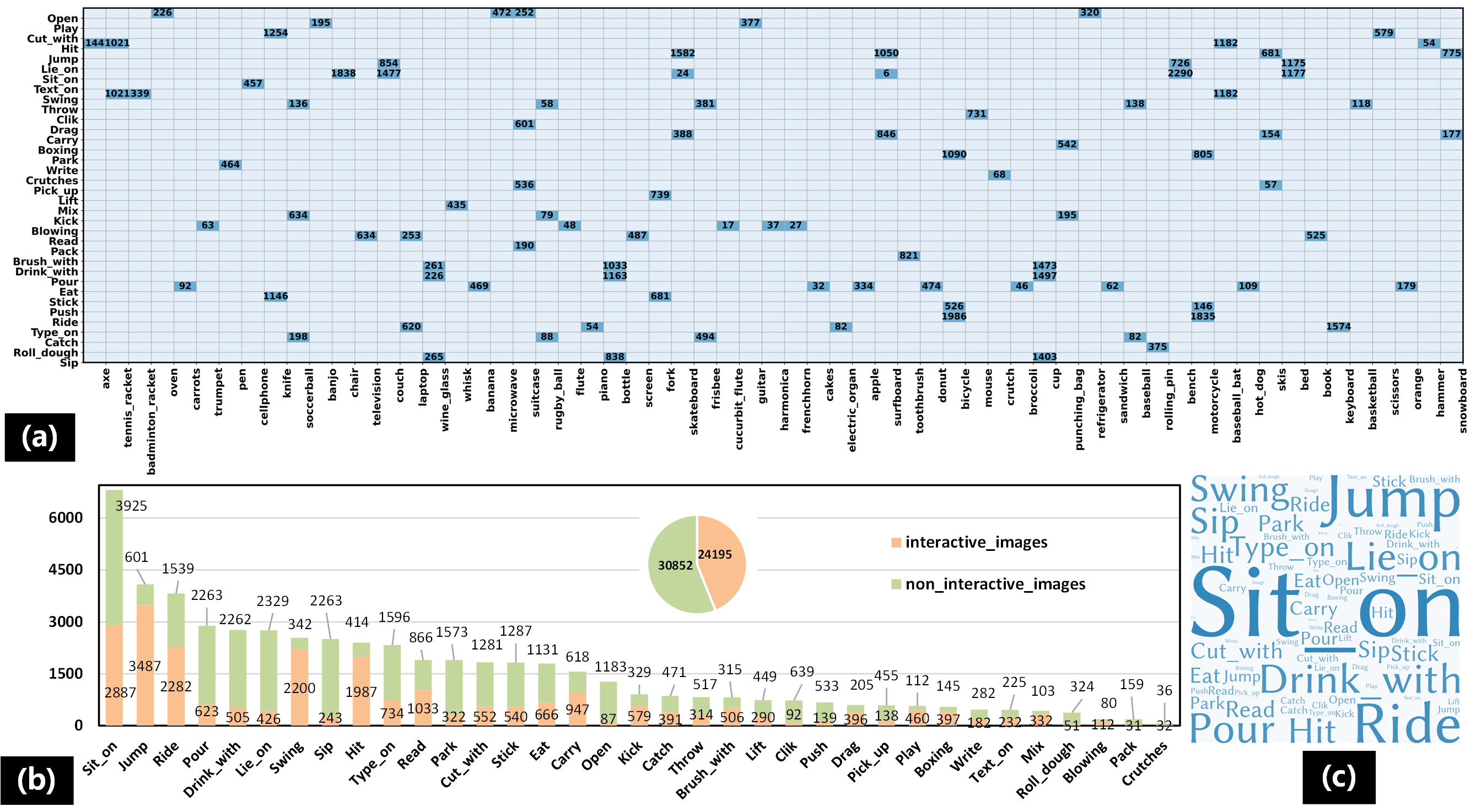}
	\end{overpic}
 \vspace{-10pt}
	\caption{\textbf{Some properties of Contact-driven Affordance Learning (CAL) dataset.} (a) Sample statistics of interactive and non-interactive images for each affordance category. (b) Affordance and object category confusion matrix, with numbers representing the samples of the current object category contained in the affordance category. (c) The word cloud distribution of affordances in the CAL dataset.  }
	\label{dataset_properties}
\end{figure*}

\subsection{Dataset Collection} 
\label{dataset collection}
In embodied intelligence applications, agents might be in a variety of scenarios and interact in a variety of complex manners. To this end, we select $35$ interaction categories that occur frequently in daily life and cover various indoor and outdoor scenes. To obtain rich human-object interaction images, we select from several datasets containing rich interactions COCO \cite{lin2014microsoft}, HICO \cite{chao2018learning} and Visual Genome (VG) \cite{krishna2017visual}. For HICO, we filter the images based on the interaction category. For COCO, we first filter the candidate images based on the category of the object and then select them based on whether or not they contain interactions. For the VG dataset, we choose candidate images based on the semantics of the objects and the relationship categories contained in the images. We select images containing more clearly defined pairs of human-object interaction relationships as part of the dataset. Further, to increase the data's diversity, we select some interactive images from PAD \cite{zhai2022one} and AGD20K \cite{luo2022learning} datasets for supplementation. Whereas for non-interactive object images, we select non-interactive objects that satisfy the semantic and affordance conditions from the scene-rich and diverse COCO \cite{lin2014microsoft}, PAD \cite{luo2021one,zhai2022one}, AGD20K \cite{luo2023grounded,luo2022learning}, and VG \cite{krishna2017visual} datasets, and we discriminate images from these datasets based on the object semantic categories and manually remove images with ambiguities. Examples of interactive and non-interactive images in the dataset are shown in Fig. \ref{dataset_example}.

\subsection{Dataset Annotation} 
\label{dataset ann}
We consider the interaction of 7 different body parts: ``Hands'', ``Feet'', ``Mouth'', ``Hips'', ``Back'', ``Eye'' and ``Outside'' (the ``Outside'' represents the region where the object contacts the outside world during the interaction, \eg, the wheel-ground contact region during riding). These categories cover almost all regions where humans interact with objects daily and thoroughly describe how various object regions are used. Since affordance learning recognizes the object's ``action possibilities'' regions, the heatmap is appropriate for describing the possibility of interactions. We refer to the previous annotation works \cite{fang2018demo2vec,bylinskii2015saliency,judd2012benchmark} and choose to annotate local regions of the image with different densities of points. We also assign object labels to the interactive and non-interactive images according to their object categories. Following Gebru et al. \cite{gebru2021datasheets}, we employ $10$ random volunteers from the laboratory, ensuring that $3$ volunteers annotated each image. For the interactive image annotation, we establish the following rules before labeling:  \textbf{(1)} According to the given interaction category, select all the interaction instances in the image that satisfy the conditions, annotate humans and objects with the bounding box, and simultaneously select the corresponding object semantic category. \textbf{(2)} For a given interaction instance, the hand (``point1''), feet (``point2''), mouth (``point3''), back (``point4''), hip (``point5''), eye (``point6''), and the corresponding region of the external interaction on the object (``point7'') are labeled. The region of interaction with the outside refers to the local region of the object that affects the outside during the interaction between the human and the object, \eg, the front half of a knife for cutting, a fork for spearing food.  \textbf{(3)} Ignore the differences between the left \& right in hands and feet.  \textbf{(4)} The points are marked more intensively for regions with frequent interaction. 

\par For the non-interactive image annotation, we establish the following rules before labeling: \textbf{(1)} Based on the examples in the interactive image, all the instances in the non-interactive image that satisfy the conditions are selected with labeling bounding boxes and corresponding object semantic categories. \textbf{(2)} For each instance of an image satisfying the conditions, the corresponding hand (``point1''), feet (``point2''), mouth (``point3''), back (``point4''), hip (``point5''), eye (``point6''), and the corresponding region of the outside interaction on the object (``point7'') are annotated according to the examples in the interactive image. \textbf{(3)} For interactive local regions on the object, annotations of various densities are provided according to the frequency of interaction.
\par In processing the annotations, we set each annotation point on the mask to $1$. Then, we utilize Gaussian blurring for each annotation point on the mask. The length and width of the image determines the size of the Gaussian kernel (ks):
\begin{equation}
\small
   ks=\frac{\sqrt{h^2+w^2}} {\sigma(k)}, \quad where \quad \sigma(k)=3,
\end{equation}
then normalize it (\ie  $(V-V_{min})/(V_{max}-V_{min}$)) to obtain the final mask. Some annotations are shown in Fig. \ref{dataset_example}.

\begin{figure}[t]
	\centering
		\begin{overpic}[width=0.97\linewidth]{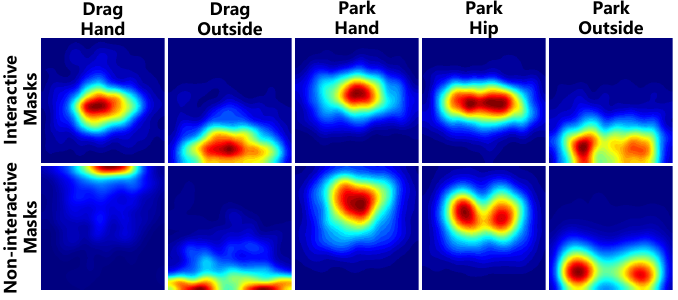}
	\end{overpic}
 \vspace{-8pt}
 \caption{\textbf{Affordance masks.} Average mask visualization of different body contact regions in the interactive and non-interactive images.}
	\label{affordance mask}
\end{figure}

\subsection{Statistic Analysis} 
\label{statistic}
To get deeper insights into the CAL dataset, we show its important features from the following aspects. Fig. \ref{dataset_properties} (a) shows the number of images in each affordance category, which shows a long-tailed distribution of the data and an uneven distribution of samples among different categories. Most categories contain comparable quantities of interactive and non-interactive images, which provides a richer way of combining them for training and increases the diversity of the samples. Fig. \ref{dataset_properties} (b) shows the confusion matrix between different affordances and objects, which presents the property of multiple possibilities of affordances, \ie, the same object may belong to more than one affordance category, and the property of multiple possibilities of one affordance category covering several different object categories. It increases the challenge of the task of affordance learning. Fig. \ref{dataset_properties} (c) exhibits the word cloud distribution of the affordance class of the CAL dataset, and it shows that we constructed the dataset to cover a wide range of affordance classes for various scenarios. We also visualized the affordance maps of different categories, as shown in Fig. \ref{affordance mask}. The masks of distinct affordance categories present diverse distribution regions, e.g., for interactive images, the location of the hand or hip contact region is mainly distributed in the middle of the image, which follows the recordings of the interaction content in daily life. On the other hand, non-interactive maps have different distribution regions from interactive maps. They are not always located in the center of the image, which makes it necessary for the model to reason about the corresponding affordance regions by using the relevant clues from the appearance and structure.

\begin{table*}[!t]
    \centering
    %\small
  \renewcommand{\arraystretch}{1.}
  \renewcommand{\tabcolsep}{10. pt}
   \caption{\textbf{The results of different methods on the CAL dataset.} The best results are in \textbf{bold}. \textbf{\texttt{Seen}} means that the training and test sets contain the same affordance/object categories, \textbf{\texttt{Obj Unseen}} denotes that the objects in the training and test sets do not overlap, and \textbf{\texttt{Aff Unseen}} represents that the affordance/objects in the training and test sets do not overlap. }
   \label{Table:total result}
   \vspace{-10pt}
  \begin{tabular}{r|ccc|ccc|ccc|c}
    \hline
    \Xhline{2.\arrayrulewidth}
  \textbf{Method}    & \multicolumn{3}{c|}{\textbf{\texttt{Seen}}} & \multicolumn{3}{c|}{\textbf{\texttt{Obj Unseen}}} & \multicolumn{3}{c|}{\textbf{\texttt{Aff Unseen}}} &  \textbf{params}   \\
  \cline{2-10}
    & $\text{KLD} \downarrow$ &  $\text{SIM} \uparrow$  & $\text{NSS} \uparrow$ & $\text{KLD} \downarrow$ &  $\text{SIM} \uparrow$  & $\text{NSS} \uparrow$  &  $\text{KLD} \downarrow$ &  $\text{SIM} \uparrow$  & $\text{NSS} \uparrow$  & (M) \\ 
    \hline
\Xhline{2.\arrayrulewidth}
 
  PSPNet \cite{zhao2017pyramid} &  $1.095$ &	$0.493$ &	$2.362$  & $1.547$ & 	$0.388$ &	$1.624$ & $1.779$ &	$0.316$ &	$1.175$ & 53.32   	\\
   DLabV3+ \cite{chen2018encoder}  & $1.034$ &	$0.533$ &	$2.605$ & $1.409$ &	$0.434$ &	$1.932$ & $1.636$ &	$0.321$ & $1.291$ & 40.35   \\
  
  SegFormer \cite{xie2021segformer}    & $1.060$ &	$0.648$ &	$3.211$ & $1.240$ & $0.539$ &	$2.533$ & $2.023$ &	$\underline{0.363}$	& $1.486$ & $27.35$     \\
    \hline
 ViTPose \cite{xu2022vitpose} &  $1.535$ &	$0.360$ &	$1.594$ & $1.715$ &	$0.330$ &	$1.239$ & $1.864$ &	$0.262$&	$0.831$ & 90.00  \\
  
  HRFormer \cite{YuanFHLZCW21}  & $0.778$ &	$0.616$ &	$3.106$ & $1.443$ &	$0.511$ & $2.353$ & $2.434$ & $0.352$ &	$1.403$ & 10.11    \\
\hline

HSNet \cite{min2021hypercorrelation}   & $1.511$ &	$0.341$ &	$1.426$ & $1.663$ &	$0.310$ &	$1.215$ & $1.774$ &	$0.279$ &	$1.130$ & 28.13  \\
\hline
CLIP \cite{radford2021learning}  & $0.883$ &	$0.538$ &	$2.944$  & $1.252$ & $0.456$ &	$2.236$ & $1.605$ &	$0.342$ &	$1.534$ & 89.88  \\
\hline

  PIANet \cite{luo2023leverage}  & $\underline{0.630}$ &	$\underline{0.680}$ &	$\underline{3.444}$ & $\underline{1.194}$ &	$\underline{0.541}$ &	$\underline{2.616}$ & $\underline{1.597}$ &	$0.362$ &	$\underline{1.601}$ & 36.32   \\

  \textbf{Ours}   & $\bm{0.595}$ & $\bm{0.692}$ &	$\bm{3.556}$  & $\bm{1.056}$ &	$\bm{0.555}$ &	$\bm{2.756}$ & $\bm{1.489}$	 & $\bm{0.382}$	& $\bm{1.711}$
 & 39.69  \\

    \hline
\Xhline{2.\arrayrulewidth}
    \end{tabular}
  \end{table*}

\section{Experiments}
\label{sec:experiments}
In this section, we first describe the experimental settings (Sec. \ref{Experimental Settings}), which include the metrics, the comparison methods, and the implementation details. Then, we compare the performance of different methods in affordance learning, both subjectively and objectively (Sec. \ref{Quantitative and Qualitative Comparisons}). Following this, we analyze the performance of the different models from various perspectives (Sec. \ref{Performance Analysis}). Finally, we conduct an ablation study of the modules in the methods to analyze the impact of different modules on the performance of affordance learning and generalization (Sec. \ref{Ablation Study}).

\subsection{Experimental Settings}
\label{Experimental Settings}
\textbf{Metrics.\ } Previous works mainly segment precise affordance regions \cite{luo2021one,chuang2018learning,myers2015affordance}, while the cross-view affordance grounding task considers a weakly supervised setting that predicts the affordance heatmap using only the affordance category label. Referring to the hotspots grounding-related works \cite{fang2018demo2vec,nagarajan2019grounded}, we adopt heatmaps to give a better description of the ``action possibilities'' (\ie, affordance) and use \textbf{KLD \cite{bylinskii2018different}}, \textbf{SIM \cite{swain1991color}}, and \textbf{NSS \cite{peters2005components}} to evaluate the probability distribution correlation between the predicted affordance heatmap and Ground Truth (GT).

\begin{figure*}[t]
	\centering
		\begin{overpic}[width=0.96\linewidth]{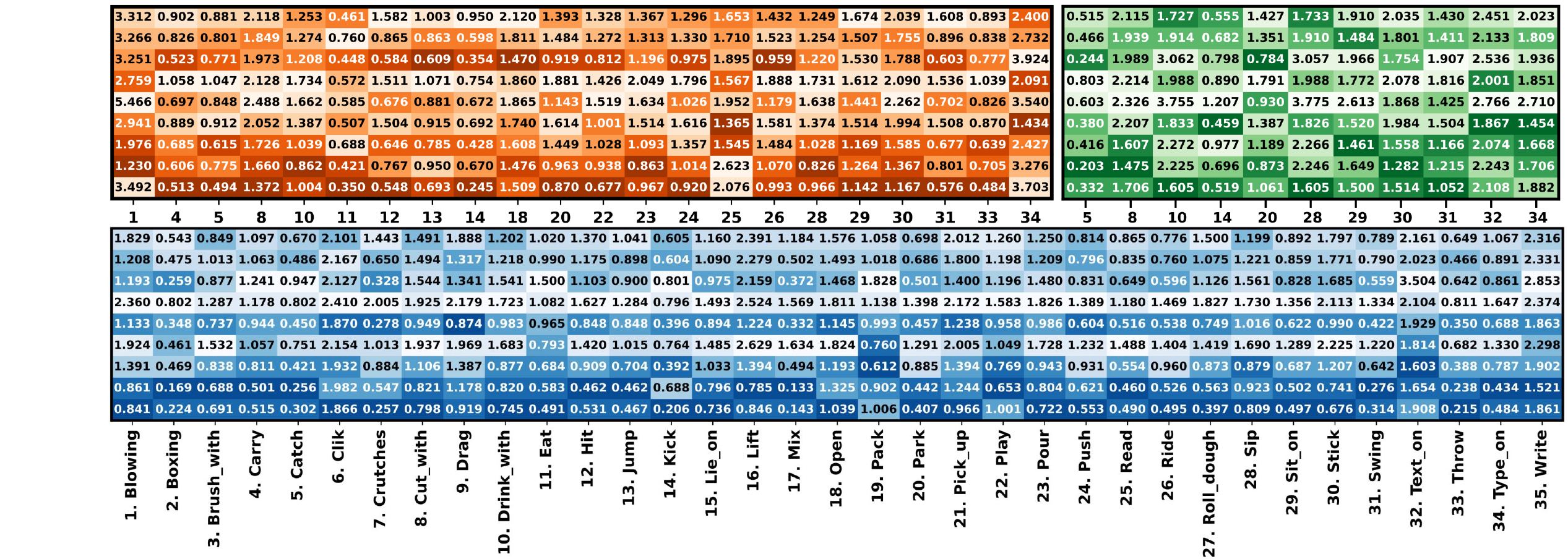}
  \put(1.35,34.4){\tiny{\textbf{PSPNet}\cite{zhao2017pyramid}}}
  \put(0.5,32.9){\tiny{\textbf{DLabV3+}\cite{chen2018encoder}}}
  \put(0.01,31.55){\tiny{\textbf{SegFormer}\cite{xie2021segformer}}}
  \put(1.4,30.2){\tiny{\textbf{VitPose}\cite{xu2022vitpose}}}
  \put(0.14,28.8){\tiny{\textbf{HRFormer}\cite{YuanFHLZCW21}}}
  \put(1.78,27.45){\tiny{\textbf{HSNet}\cite{min2021hypercorrelation}}}
  \put(2.57,26.1){\tiny{\textbf{CLIP}\cite{radford2021learning}}}
  \put(1.5,24.7){\tiny{\textbf{PIANet}\cite{luo2023leverage}}}
  \put(4.5,23.4){\tiny{\textbf{Ours}}}

  \put(1.35,20.2){\tiny{\textbf{PSPNet}\cite{zhao2017pyramid}}}
  \put(0.5,18.7){\tiny{\textbf{DLabV3+}\cite{chen2018encoder}}}
  \put(0.01,17.3){\tiny{\textbf{SegFormer}\cite{xie2021segformer}}}
  \put(1.4,15.95){\tiny{\textbf{VitPose}\cite{xu2022vitpose}}}
  \put(0.14,14.55){\tiny{\textbf{HRFormer}\cite{YuanFHLZCW21}}}
  \put(1.78,13.2){\tiny{\textbf{HSNet}\cite{min2021hypercorrelation}}}
  \put(2.57,11.9){\tiny{\textbf{CLIP}\cite{radford2021learning}}}
  \put(1.5,10.55){\tiny{\textbf{PIANet}\cite{luo2023leverage}}}
  \put(4.5,9.2){\tiny{\textbf{Ours}}}
	\end{overpic}
 \vspace{-8pt}
	\caption{\textbf{Different Classes.}  We measure the KLD, SIM, and NSS metrics for each affordance category, with darker colors representing higher performance. The \textcolor[rgb]{0,0.5,1}{\textbf{blue}}, \textcolor[rgb]{1,0.5,0}{\textbf{orange}} and \textcolor[rgb]{0.2,0.6,0}{\textbf{green}}  colours represent \textbf{\texttt{Seen}}, \textbf{\texttt{Obj Unseen}} and \textbf{\texttt{Aff Unseen}}, respectively.}
	\label{Diff classes}
\end{figure*}

\textbf{Comparison methods.\ } In order to more thoroughly evaluate the advantages of our model on the affordance learning task, we compare representative models from multiple domains. Since this paper's task and segmentation model assign corresponding labels based on the image content, we select advanced models in the segmentation field for comparison (\textbf{PSPNet} \cite{zhao2017pyramid}, \textbf{DLabV3+} \cite{chen2018encoder} and \textbf{SegFormer} \cite{xie2021segformer}). Because our work needs to mine the interactive affinity from human-object interactions, we chose the classical human pose estimation models (\textbf{VitPose} \cite{xu2022vitpose} and \textbf{HRFormer} \cite{YuanFHLZCW21}) to evaluate the model's ability to perceive the interactive pose. On the other hand, the task setting of this paper is related to few-shot segmentation and text-guided image segmentation. Hence, we choose representative few-shot segmentation (\textbf{HSNet} \cite{min2021hypercorrelation}) and multimodal models (\textbf{CLIP} \cite{radford2021learning}) for comparison. Finally, we also compare the PIANet model \cite{luo2023leverage} in the conference version to evaluate the performance of the affordance learning approach.

\textbf{Implementation details.\ } Our model is implemented in PyTorch and trained with the AdamW \cite{loshchilov2017decoupled} optimizer. With random horizontal flipping, the input images are cropped to 224$\times$224. We train the model for 35 epochs on a single NVIDIA RTX3090 GPU with an initial learning rate of 6$e$-5. We divide the dataset into \textbf{\texttt{Seen}}, \textbf{\texttt{Obj Unseen}}, and \textbf{\texttt{Aff Unseen}} settings, where the \textbf{\texttt{Seen}} setting splits the dataset into training, test, and validation sets according to 7:2:1. For the \textbf{\texttt{Obj Unseen}} setting, we select 54 categories of objects as the training set and the remaining 24 categories as the testing and validation set, followed by dividing this sample into the testing and validation sets according to 2:1, which is mainly used for evaluating the model's ability to understand and generalize the interactable regions of unseen objects. For the \textbf{\texttt{Aff Unseen}} setting, we select 24 of the affordance categories as the training set. The remaining portion is used as the testing and validation, dividing the testing and validation sets according to 2:1, mainly employed to assess the model's generalization ability for unseen objects with unseen interactions. In \textbf{\texttt{Seen}} and \textbf{\texttt{Obj Unseen}}  settings, we set the batch size to 24 and train for 120,000 iterations, whereas in \textbf{\texttt{Aff Unseen}}, we take the batch size to 12 and train for 20,000 iterations.

\begin{figure*}[t]
	\centering
		\begin{overpic}[width=0.95\linewidth]{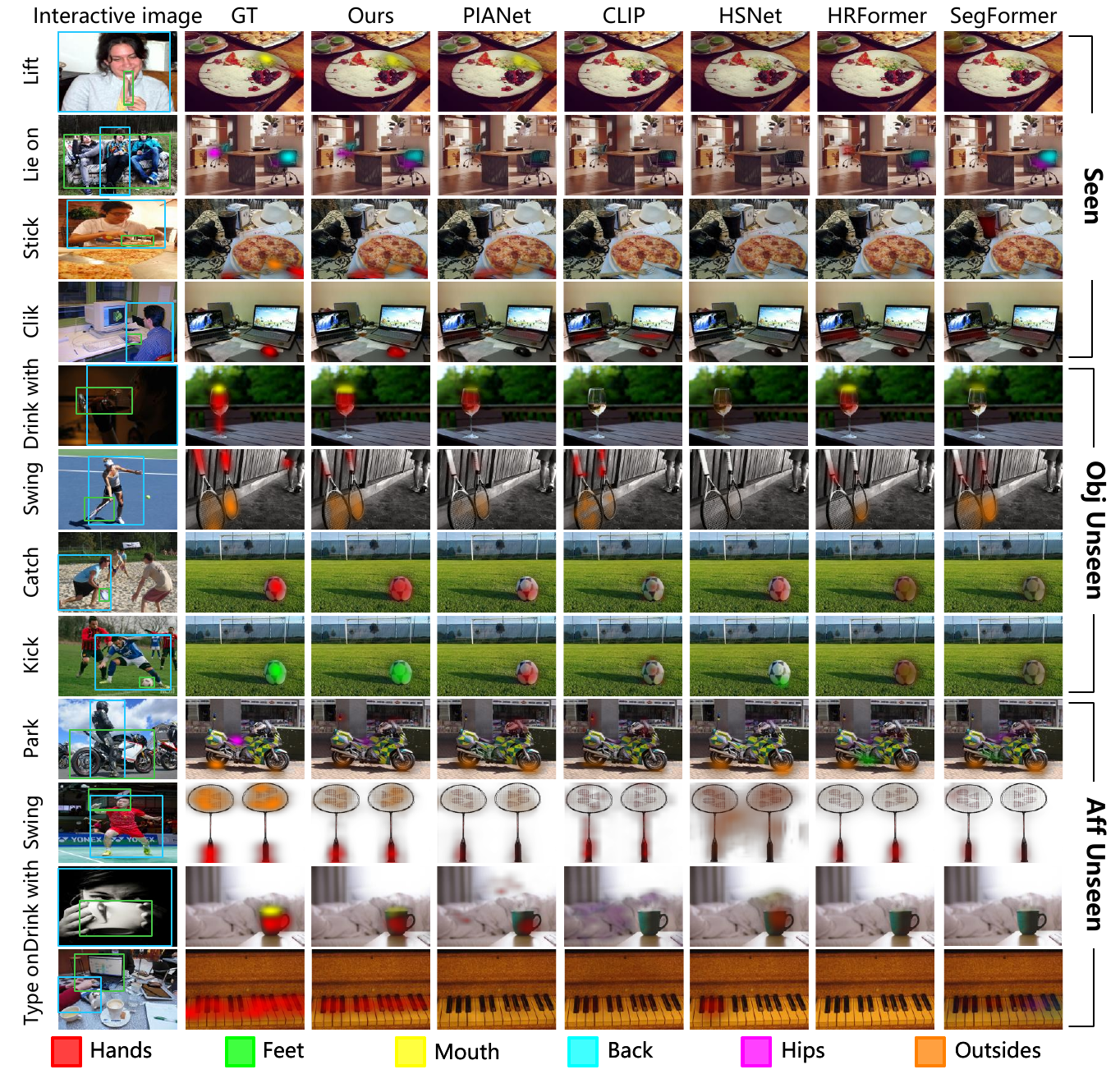}
	\end{overpic}
 \vspace{-10pt}
	\caption{\textbf{Visualization of prediction affordance maps.} We show the visualization results of our model, affordance learning model (PIANet \cite{luo2023leverage}), multimodal model (CLIP \cite{radford2021learning}), few-shot segmentation model (HSNet \cite{min2021hypercorrelation}), the best human pose estimation  model (HRFormer \cite{YuanFHLZCW21}) and the segmentation model (SegFormer \cite{xie2021segformer}). }
	\label{viz maps}
\end{figure*}

\subsection{Quantitative and Qualitative Comparisons}
\label{Quantitative and Qualitative Comparisons}
The experimental results are shown in Table \ref{Table:total result}. Our approach achieves the best performance on multiple metrics across all settings, taking the KLD metrics as an example, with the \textbf{\texttt{Seen}} setting, our method improves \textbf{43.87\%} compared to the best segmentation model, increases \textbf{23.52\%} compared to the best human pose estimation method, gains \textbf{60.62\%} relatively compared to the few-shot segmentation approach, exceeds the multimodal model by \textbf{32.61\%}, and improves \textbf{5.56\%} compared to the affordance learning's model. In the \textbf{\texttt{Obj Unseen}} setting, our model outperforms the advanced segmentation model by \textbf{14.84\%}, superior to the best human pose estimation method by \textbf{26.82\%}, higher than the network of few-shots by \textbf{36.50\%}, with a relative improvement of \textbf{15.66\%} compared to the multimodal model, and outperforms PIANet by \textbf{11.56\%}. In the Aff Unseen setting, our model exceeds the advanced segmentation model by \textbf{26.40\%}, excels the best human pose estimation method by \textbf{38.83\%}, outperforms the few-shot segmentation network by \textbf{16.07\%}, and has a relative improvement of \textbf{7.23\%} compared to the multimodal model, and outperforms PIANet by \textbf{6.76\%}.  Meanwhile, we visualize the affordance prediction results of the different methods, as shown in Fig. \ref{viz maps}. Our method accurately predicts the affordance region of an object in all three settings, \textbf{\texttt{Seen}}, \textbf{\texttt{Obj Unseen}}, and \textbf{\texttt{Aff Unseen}}. Specifically, our approach works better to locate interactive object localised regions in non-interactive images without activating other irrelevant object regions in the interactive image scene (\eg, the fourth row). For cases where the scene contains multiple objects (\eg, the second, third, sixth, and tenth rows), the model is also capable of positioning all of them, suggesting that our consideration of pose as a cue between interactive and non-interactive images can help the model to more accurately perceive candidate regions based on the structure of the objects. For cases where the same object may belong to more than one affordance category (\eg, rows 7 and 8), our method can also predict the corresponding affordance maps based on different body-contact regions, eliminating the ambiguity caused by the multiplicity of possibilities of affordance.

\begin{table*}[!t]
    \centering
    %\small
    %\footnotesize
    \scriptsize
  \renewcommand{\arraystretch}{1.}
  \renewcommand{\tabcolsep}{2.5pt}
   \caption{\textbf{Different parts.} We evaluate the predictions for each body part and object contact region. The \textbf{Bold} and \textbf{\underline{Underline}} represent the best and sub-optimal results, respectively.}
   \label{Table:different parts}
   \vspace{-10pt}
  \begin{tabular}{c|c|ccc|ccc|ccc|ccc|ccc|ccc|ccc}
    \hline
    \Xhline{2.\arrayrulewidth}
        & \textbf{Method}  & \multicolumn{3}{c|}{\textbf{Hand}} & \multicolumn{3}{c|}{\textbf{Feet}} & \multicolumn{3}{c|}{\textbf{Mouth}} & \multicolumn{3}{c|}{\textbf{Hips}} & \multicolumn{3}{c|}{\textbf{Back}} &  \multicolumn{3}{c|}{\textbf{Eyes}} & \multicolumn{3}{c}{\textbf{Outsides}}  \\ 
        \Xhline{2.\arrayrulewidth}
        \multirow{6}{*}{\rotatebox{90}{\textbf{\texttt{Seen}}}} &  
          SegFormer &  1.263 &	0.629 &	3.002 &	0.923 &	0.610 &	3.240 &	1.442 &	0.633 &	3.247 &	0.929 &	0.609 &	2.663 &	0.703 &	\color{blue}{0.677} &	3.980 &	1.136 &	0.680 &	2.719 &	0.898 &	0.691 &	3.278   \\
        
        & HFormer &  0.859 &	0.593 &	2.904 &	0.768 &	0.575 &	3.036 &	1.028 &	0.610 &	3.238 &	0.748 &	0.578 &	2.555 &	0.599 &	0.646 &	3.828 &	0.785 &	0.677 &	2.748 &	0.673 &	0.658 &	3.175  \\
        & HSNet &  1.511 &	0.341 &	1.426 &	1.367 &	0.384 &	1.771 &	1.800 &	0.268 &	1.184 &	1.277 &	0.377 &	1.601 &	1.460 &	0.324 &	1.795 &	1.500 &	0.351 &	1.162 &	1.254 &	0.424 &	1.887   \\
       % \rowcolor{mygray}
         & CLIP & 0.935 &	0.524 &	2.806 &	0.896 &	0.509 &	2.790 &	0.971 &	0.529 &	3.234 &	0.821 &	0.547 &	2.524 &	0.855 &	0.536 &	3.438 &	0.690 &	0.645 &	2.844 &	0.818 &	0.562 &	2.950 \\
        \cline{2-23}
      &  PIANet & \color{blue}{0.669} &	\color{blue}{0.673} &	\color{blue}{3.320} &	\color{blue}{0.607} &	\color{red}{\textbf{0.657}} &	
      \color{blue}{3.564} &	\color{blue}{0.865} &	\color{blue}{0.656} &	\color{blue}{3.465} &	\color{blue}{0.600} &	\color{blue}{0.660} &	\color{blue}{2.966} &	\color{blue}{0.548} &	\color{blue}{0.677} &	\color{blue}{3.988} &	\color{blue}{0.684} &	\color{blue}{0.697} &	\color{blue}{2.914} &	\color{blue}{0.541} &	\color{blue}{0.717} &	\color{blue}{3.473}  \\
         & Ours &  \color{red}{\textbf{0.663}} &	\color{red}{\textbf{0.677}} &	\color{red}{\textbf{3.400}} &	\color{red}{\textbf{0.588}} &	\color{blue}{0.656} &	\color{red}{\textbf{3.589}} &	\color{red}{\textbf{0.799}} &	\color{red}{\textbf{0.673}} &	\color{red}{\textbf{3.615}} &	\color{red}{\textbf{0.548}} &	\color{red}{\textbf{0.689}} &	\color{red}{\textbf{3.118}} &	\color{red}{\textbf{0.496}} &	\color{red}{\textbf{0.694}} &	\color{red}{\textbf{4.015}} &	\color{red}{\textbf{0.657}} &	\color{red}{\textbf{0.705}} &	\color{red}{\textbf{2.930}} &	\color{red}{\textbf{0.497}} &	\color{red}{\textbf{0.722}} &	\color{red}{\textbf{3.525}}  \\
        \hline
    \Xhline{2.\arrayrulewidth}
        \multirow{6}{*}{\rotatebox{90}{\textbf{\texttt{Obj Unseen}}}}
        &  SegFormer & 1.490 & 0.504 &	2.361 &	\color{blue}{0.889} &	\color{red}{0.551} &	\color{blue}{2.992} &	\color{blue}{1.261} &	\color{blue}{0.557} &	3.072 &	1.507 &	0.461 &	1.844 &	1.313 &	0.484 &	3.183 &	\color{red}{0.620} &	\color{red}{0.625}	& \color{red}{1.957} &	0.935 &	0.620 &	2.394  \\
       
        & HFormer &  1.671 &	0.490 &	2.237 &	1.074 &	0.509 &	2.754 &	1.759 &	0.508 &	2.812 &	1.415 &	0.464 &	1.929 &	1.570 &	0.422 &	2.607 &	1.064 &	\color{blue}{0.578} &	1.659 &	1.147 &	0.590 &	2.306  \\
        & HSNet &  1.663 &	0.310 &	1.215 &	1.437 &	0.354 &	1.873 &	1.755 &	0.268 &	1.031 &	1.436 &	0.334 &	1.351 &	1.697 &	0.273 &	1.834 &	1.081 &	0.428 &	1.136 &	1.205 &	0.431 &	1.564 \\
        & CLIP & 1.398 &	0.431 &	2.097 &	1.180 &	0.442 &	2.458 &	1.097 &	0.492 &	2.859 &	\color{blue}{1.112} &	0.478 &	2.125 &	1.601 &	0.366 &	2.602 &	\color{blue}{0.768} &	0.566 &	\color{blue}{1.912} &	0.972 &	0.524 &	2.133  \\
        \cline{2-23}
        & PIANet & \color{blue}{1.388} &	\color{blue}{0.507} &	\color{blue}{2.423} &	1.245 &	0.508 &	2.730 &	\color{red}{\textbf{0.893}} &	\color{red}{\textbf{0.567}} &	\color{red}{\textbf{3.151}} &	1.238 &	\color{blue}{0.520} &	
        \color{blue}{2.223} &	\color{blue}{1.173} &	\color{red}{\textbf{0.509}} &	\color{red}{\textbf{3.430}} &	2.080 &	0.428 &	1.186 &	\color{blue}{0.877} &	\color{blue}{0.627} &	\color{blue}{2.558} \\
         & Ours &  \color{red}{\textbf{1.262}} &	\color{red}{\textbf{0.513}} &	\color{red}{\textbf{2.542}} &	\color{red}{\textbf{0.885}} &	\color{blue}{0.548} &	\color{red}{\textbf{3.158}}&	1.131 &	0.534 &	3.019 &	\color{red}{\textbf{1.046}} &	\color{red}{\textbf{0.528}} &	\color{red}{\textbf{2.387}} &	\color{red}{\textbf{1.121}} &	\color{blue}{0.492} &	\color{blue}{3.331} &	0.980 &	0.517 &	1.414 &	\color{red}{\textbf{0.774}} &	\color{red}{\textbf{0.635}} &	\color{red}{\textbf{2.707}} \\
         \hline
    \Xhline{2.\arrayrulewidth}
    \multirow{6}{*}{\rotatebox{90}{\textbf{\texttt{Aff Unseen}}}} 
        &  SegFormer & 2.243 &	0.310 &	1.317 &	0.798 &	0.568 &	1.425 &	2.788 &	
        \color{blue}{0.262} &	0.998 &	2.230 &	0.343 &	\color{blue}{1.072} &	1.548 &	\color{red}{0.441} &	\color{blue}{2.227} &	2.813 &	0.308 &	0.881 &	1.484 &	0.458 &	1.845    \\
       
        & HFormer & 2.429 &	0.329 &	1.409 &	1.207 &	0.519 & 	1.372 &	3.988 &	0.215 &	0.692 &	2.958 &	0.292 &	0.846 &	2.052 &	0.399 &	1.906 &	3.277 &	0.294 &	0.895 &	1.631 &	0.462 &	1.804  \\
        & HSNet & 1.774 &	0.279 &	1.130 &	\color{red}{\textbf{0.459}} &	\color{blue}{0.673} &	\color{red}{\textbf{2.079}} &	\color{blue}{1.971} &	0.229 &	0.785 &	\color{red}{\textbf{1.464}} &	0.334 &	\color{red}{\textbf{1.210}} &	1.601 &	0.293 &	1.326 &	\color{red}{\textbf{1.765}} &	0.288 & 	\color{blue}{1.171} &	1.431 &	0.371 &	1.361 \\
       & CLIP  & 1.720	&  0.303 &	1.541 &	0.977 &	0.505 &	1.496 &	2.341 &	0.217 &	0.694 &	\color{blue}{1.582} &	\color{red}{\textbf{0.350}} &	1.032 &	\color{blue}{1.355} &	0.380 &	2.068 &	\color{blue}{2.321} &	\color{red}{\textbf{0.340}} &	\color{red}{\textbf{1.342}} &	1.189 &	0.435 &	1.904 \\
        \cline{2-23}
        & PIANet & \color{blue}{1.618} &  \color{red}{\textbf{0.357}} & \color{blue}{1.719} &	0.696 &	0.578 &	\color{blue}{2.043} &	2.281 &	0.245 &	\color{blue}{1.075} &	1.654 &	0.320 &	0.965 &	1.530  &	0.346 &	1.742 &	2.625 &	0.286 &	1.047 &	\color{blue}{1.166} &	
        \color{blue}{0.468} &	\color{blue}{2.004} \\
         & Ours & \color{red}{\textbf{1.607}} &	\color{blue}{0.343} &	\color{red}{\textbf{1.726}}	&  \color{blue}{0.556} &	\color{blue}{0.636} &	1.768 &  \color{red}{\textbf{1.924}} &	\color{red}{\textbf{0.292}} &	\color{red}{\textbf{1.435}} &	1.679 &	 \color{blue}{0.344} &	1.028 &	\color{red}{\textbf{1.337}} &	\color{blue}{0.424} &	\color{red}{\textbf{2.291}} &	2.551 &	\color{blue}{0.309} &	1.103 &	 \color{red}{\textbf{1.143}} &	\color{red}{\textbf{0.479}} &	\color{red}{\textbf{2.024}} \\
         \hline
    \Xhline{2.\arrayrulewidth}
    \end{tabular}
  \end{table*}

  \begin{figure}[t]
	\centering
		\begin{overpic}[width=0.93\linewidth]{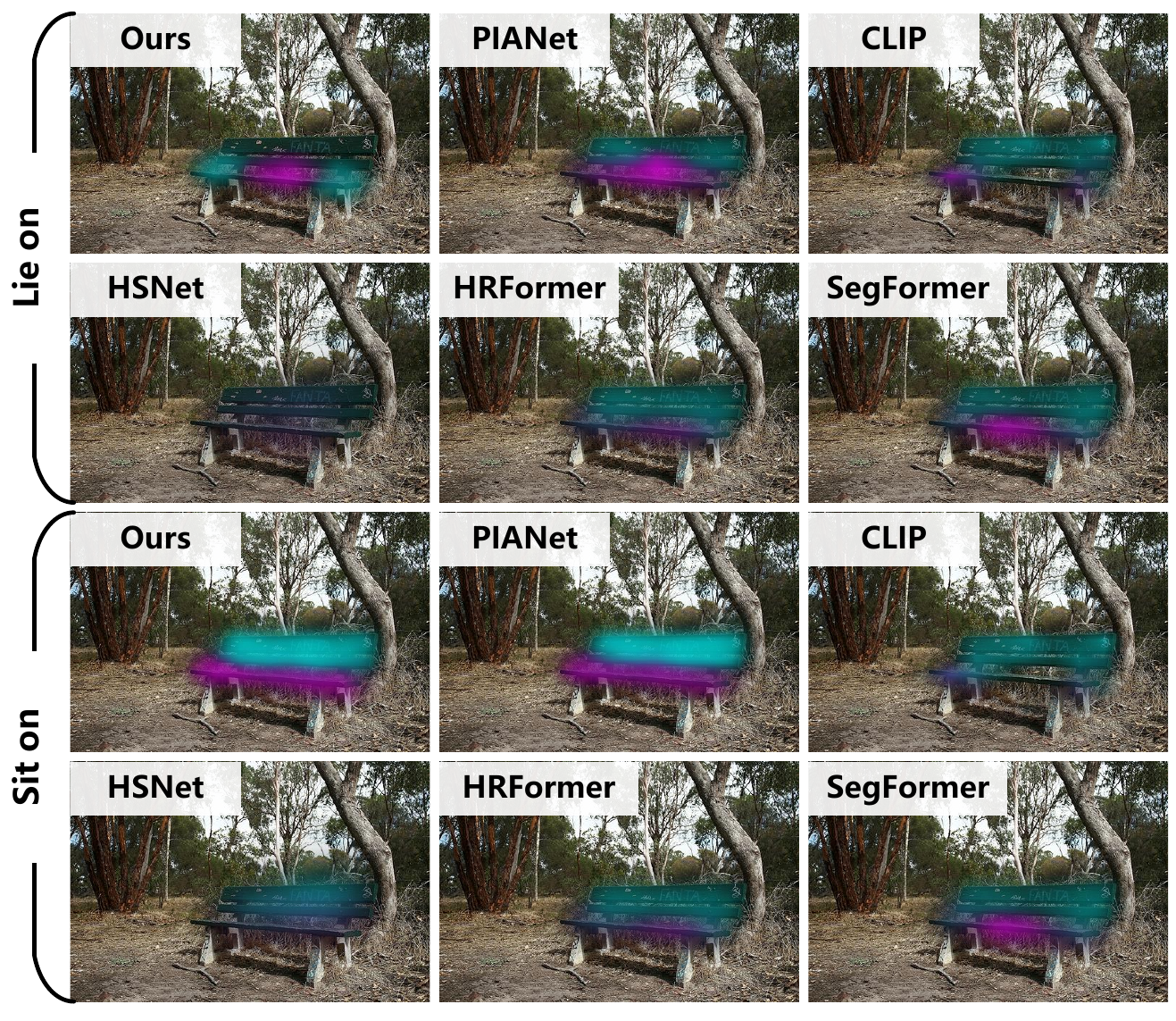}
	\end{overpic}
 \vspace{-10pt}
	\caption{\textbf{Different interactive images} \wrt  \textbf{Same non-interactive images.} We show the results for predicting the back contact region under the interaction of ``Sit on'' and ``Lie on''. }
	\label{diff_aff}
\end{figure}

\subsection{Performance Analysis}
\label{Performance Analysis}
\textbf{Different classes.\ } To evaluate the perceived ability in different categories, we show the KLD metrics for each model in each affordance category, as shown in Fig. \ref{Diff classes}, where darker colours represent a higher ranking of the model's performance under the current category. Our approach is basically the best result in \textbf{\texttt{Seen}}, and in \textbf{\texttt{Obj Unseen}}, it still achieves better performance for ``Park'' and ``Push'', which involve varied and complex scenes with large differences in the appearance of the objects, which suggests that our approach can achieve a more accurate perception of unseen objects based on the appearance and structural similarity of the objects. Our method achieves competitive performance in most of categories in \textbf{\texttt{Aff Unseen}} setting, and in the ``Drink with'' and ``Sip'' categories involving bottles, cups, and wine glasses, the performance of our method is more impressive than other methods, which indicates that our model can suppress irrelevant backgrounds in the scene by mining the interactive affinity.

\begin{figure}[t]
	\centering
		\begin{overpic}[width=0.935\linewidth]{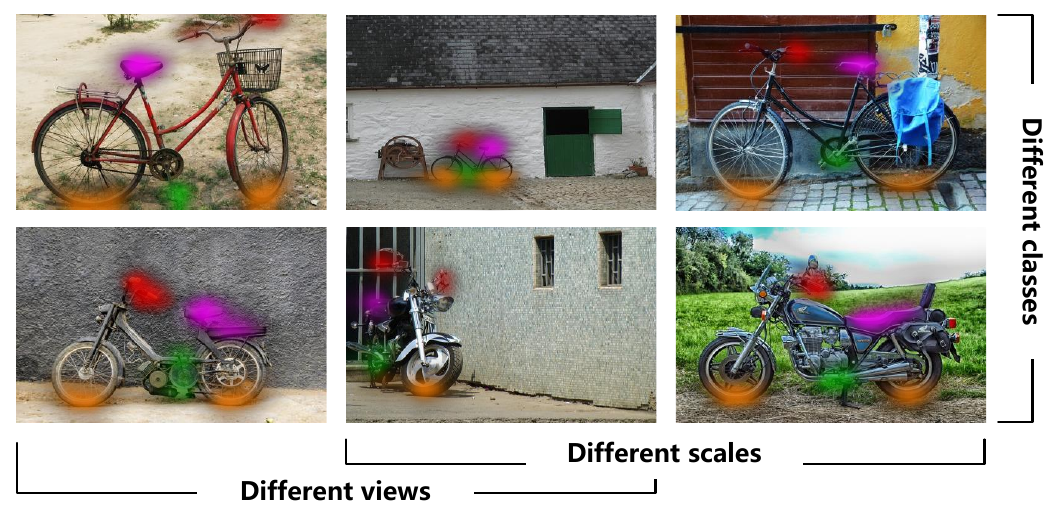}
	\end{overpic}
 \vspace{-10pt}
	\caption{\textbf{Different interactive images} \wrt  \textbf{Same non-interactive images.} We present the prediction results of the model for different non-interactive images of the same interactive image. }
	\label{diff_img}
\end{figure}

\textbf{Different body parts.\ }We tested the prediction results for different body parts to study the model's capability to perceive the interaction regions corresponding to different body parts, as shown in Table \ref{Table:different parts}. Our model achieves almost the best results at the \textbf{\texttt{Seen}} setting. Our method still maintains optimal or sub-optimal results on most metrics at the \textbf{\texttt{Obj Unseen}} setting. Under the setting of \textbf{\texttt{Aff Unseen}}, our model retains very competitive performance on most metrics, among which, for cases like ``Mouth'' where the contact region is small and occluded, our model has a considerable advantage over other methods, which may be because we consider the use of a large language model to guide the network to explicitly focus on the contact regions of different body parts, eliminating the multiple body parts' ambiguity triggered when they act on the object at the same time, thus achieving more accurate predictions.

\begin{figure}[t]
	\centering
		\begin{overpic}[width=0.95\linewidth]{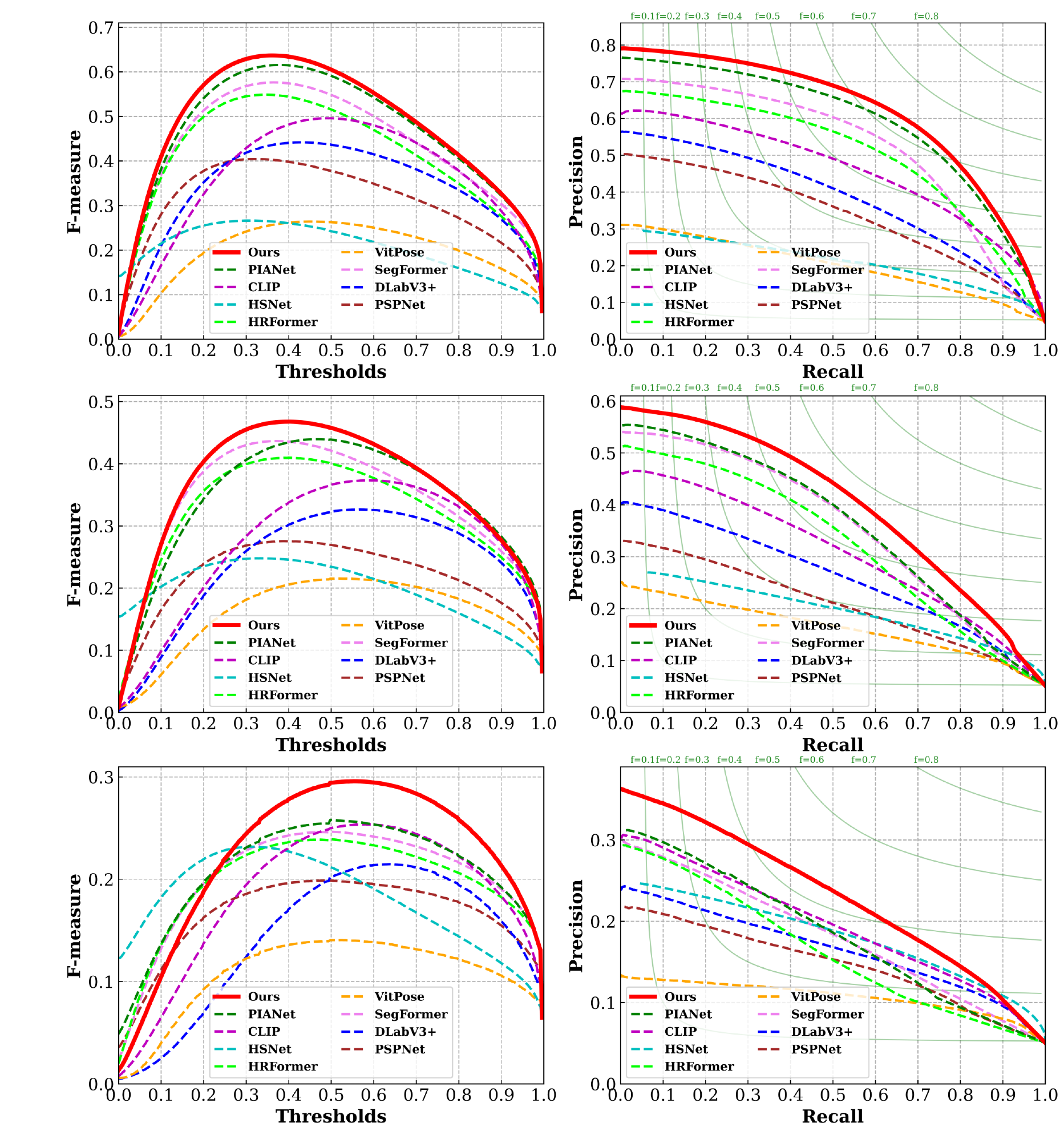}
  \put(0.55,78.9){\rotatebox{90}{{\color{black} \small{\textbf{\texttt{Seen}}}}}}
   \put(0.55,39.6){\rotatebox{90}{{\color{black} \small{\textbf{\texttt{Obj Unseen}}}}}}
    \put(0.55,5.8){\rotatebox{90}{{\color{black} \small{\textbf{\texttt{Aff Unseen}}}}}}
	\end{overpic}
 \vspace{-10pt}
	\caption{\textbf{F-measure curves and PR curves of $9$ models on the CAL dataset.} The left and right are the F-curve and PR-curve under different settings, respectively. }
	\label{F-curve and P-R-curve}
\end{figure}

\textbf{Affordance's ambiguous uncertainty.\ }  Fig. \ref{diff_aff} shows the results of different interaction images for the same non-interactive image. Since ``Sit on'' and ``Lie on'' correspond to different contact regions on the back, we only visualize this part. Our model can perceive interaction variations and accurately transfer the interaction-related invariant features to the corresponding regions. Fig. \ref{diff_img} shows the prediction for the same interactive image corresponding to different non-interactive images. Although different object categories, various scales, and different views, our model can transfer the interactive affinity correctly for each part, which indicates that the interactive affinity extraction  module can effectively establish the connection between the interactable regions from different branches and facilitate the transfer of interactive affinity.

\begin{table}[!t]
    \centering
    %\small
    \scriptsize
  \renewcommand{\arraystretch}{1.}
  \renewcommand{\tabcolsep}{1.1pt}
   \caption{\textbf{Ablation study.} We compare the impact of text guidance, pose guidance, and feature similarity comparisons on model performance. }
   \label{Table:ablation result}
   \vspace{-8pt}
  \begin{tabular}{c|ccc|ccc|ccc}
    \hline
    \Xhline{2.\arrayrulewidth}
        & \multicolumn{3}{c|}{\textbf{\texttt{Seen}}} & \multicolumn{3}{c|}{\textbf{\texttt{Obj Unseen}}} & \multicolumn{3}{c}{\textbf{\texttt{Aff Unseen}}} \\
    \cline{2-10}
    %\Xhline{2.\arrayrulewidth}
    & $\text{KLD} \downarrow$ &  $\text{SIM} \uparrow$  & $\text{NSS} \uparrow$ & $\text{KLD} \downarrow$ &  $\text{SIM} \uparrow$  & $\text{NSS} \uparrow$ & $\text{KLD} \downarrow$ &  $\text{SIM} \uparrow$  & $\text{NSS} \uparrow$ \\ 
    \hline
\Xhline{2.\arrayrulewidth}
  w/o text  & $0.613$ &	$0.687$ &	$3.518$ & $1.259$ &	$0.540$	 & $2.629$ & $1.641$ &	$0.365$ &	$1.599$\\
  w/o pose & $0.664$ &	$0.683$ &	$3.458$ & $1.219$ &	$0.544$ &	$2.625$ & $1.635$ &	$0.366$ &	$1.615$ \\
w/o app & $0.708$ &	$0.665$ & $3.385$ & $1.252$ & 	$0.532$ &	$2.565$ & $1.624$ &	$0.351$ &	$1.548$ \\

\hline
Ours & $\bm{0.595}$ & $\bm{0.692}$ &	$\bm{3.556}$  & $\bm{1.056}$ &	$\bm{0.555}$ &	$\bm{2.756}$ & $\bm{1.489}$	 & $\bm{0.382}$	& $\bm{1.711}$ \\
    \hline
\Xhline{2.\arrayrulewidth}
    \end{tabular}
  \end{table}

  \begin{figure}[t]
	\centering
		\begin{overpic}[width=0.95\linewidth]{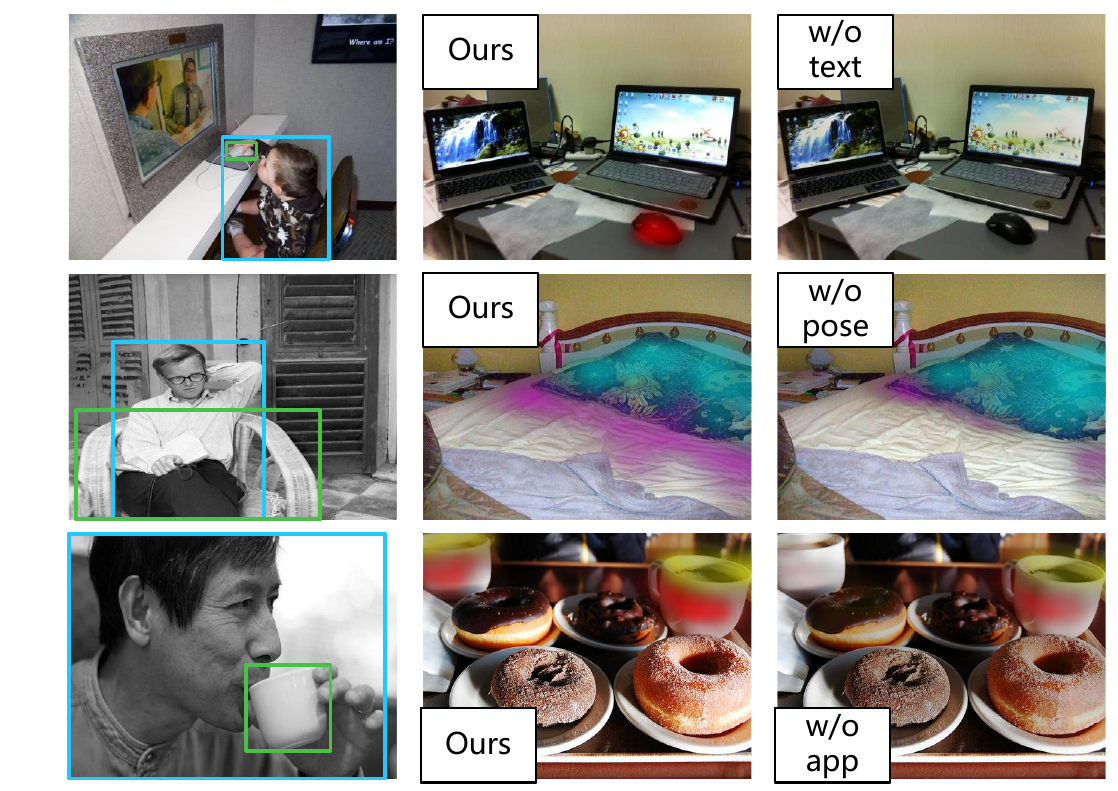}
 
  \put(0.3,57){\textbf{(a)}}
  \put(0.3,35){\textbf{(b)}}
  \put(0.3,10){\textbf{(c)}}
	\end{overpic}
 \vspace{-10pt}
	\caption{\textbf{Visualization results for ablation study.} We visualize the results of ablation study for the text, pose and apparent similarity.}
 \label{viz_ablation}
\end{figure}

  \begin{figure}[t]
	\centering
		\begin{overpic}[width=0.95\linewidth]{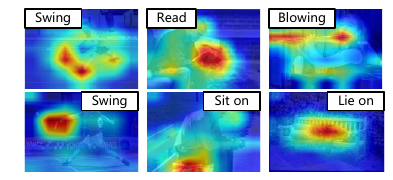}
  \put(0.2,30){\textbf{(a)}}
  \put(0.2,10){\textbf{(b)}}
	\end{overpic}
 \vspace{-10pt}
	\caption{\textbf{Visualization for semantic and pose guidance of interaction regions.} (a) Effect of semantics on interactive feature in the SPH module. (b) Effect of pose on interactive feature in the SPH module }
 \label{viz_shp}
\end{figure}

\begin{figure}[t]
	\centering
		\begin{overpic}[width=0.95\linewidth]{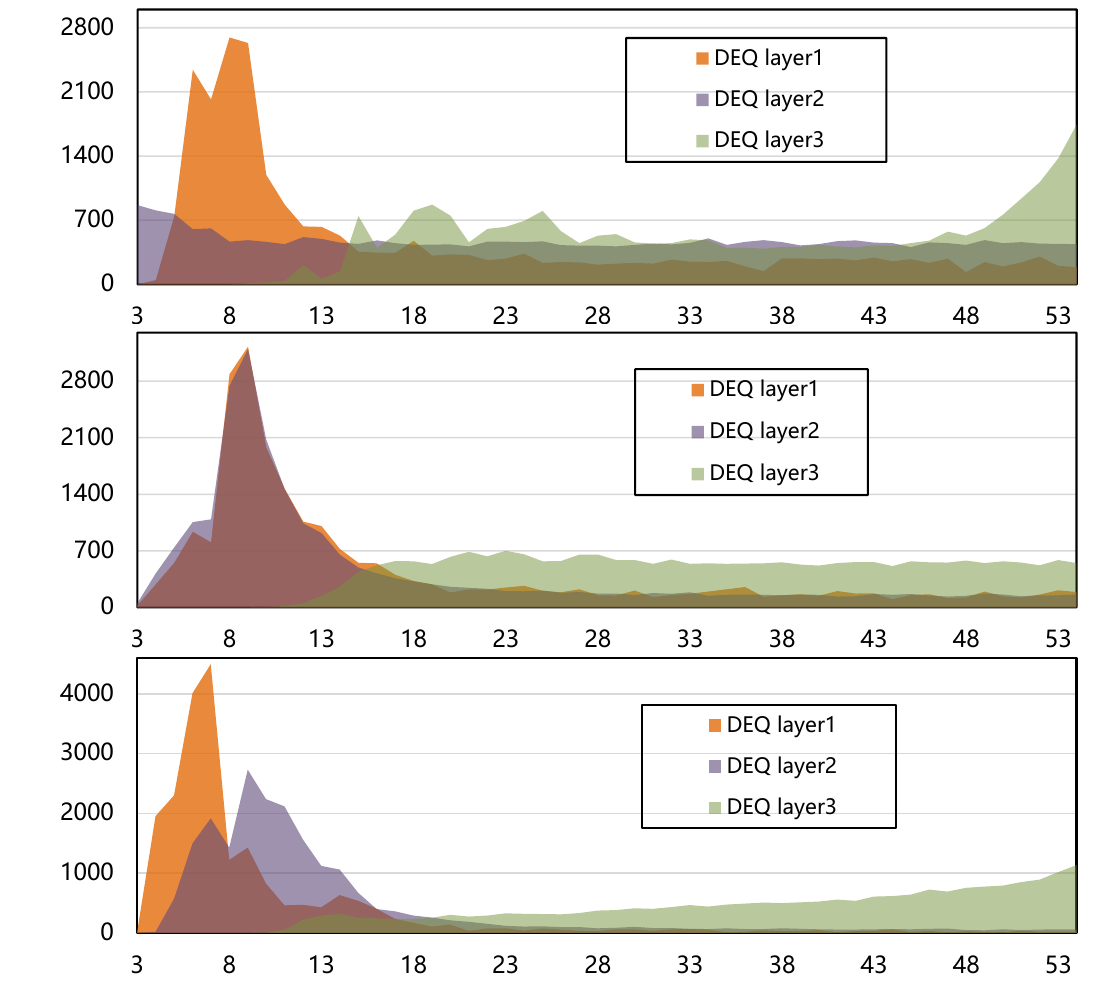}
  \put(0.6,70.6){\rotatebox{90}{{\color{black} \small{\textbf{\texttt{Seen}}}}}}
   \put(0.6,35.6){\rotatebox{90}{{\color{black} \small{\textbf{\texttt{Obj Unseen}}}}}}
    \put(0.6,4.9){\rotatebox{90}{{\color{black} \small{\textbf{\texttt{Aff Unseen}}}}}}

	\end{overpic}
 \vspace{-10pt}
	\caption{\textbf{Number of DEQ fusion layer iterations.} These represent the results of the statistical distribution of the number of iterations under different settings, respectively.}
	\label{deq1}
\end{figure}

\textbf{F-curve and P-R-curve.\ } The F-measure and PR curves are shown in  Fig. \ref{F-curve and P-R-curve}.  Our method's PR and F-measure curve are at the top in most situations, indicating that it achieves the best performance on average.  And it is even more pronounced in the \textbf{\texttt{Aff Unseen}} setting, which suggests that our method has a greater advantage than other approaches when dealing with unseen interactions and objects.

\subsection{Ablation Study}
\label{Ablation Study}
In order to verify the effectiveness of our approach to jointly mine interactive affinity from both semantic as well as structural similarity and the necessity of transferring affinity cues based on geometric and apparent similarity, we conducted the corresponding ablation experiments and visualized some of the predictions, as shown in Table \ref{Table:ablation result} and Fig. \ref{viz_ablation}. It shows that text has a more negligible effect on model performance, while apparent similarity has a larger influence on the model, suggesting a larger relationship between the possible interactions exhibited by similar local regions. From the visualization results, it is evident that text enables the model to more accurately localize the region of the object that the person is interacting with (\eg, Fig. \ref{viz_ablation} (a)). The human pose can assist the network to more accurately perceive the relationship between the object structure and the body, and accurately reason about the interactable regions corresponding to different objects (\eg, Fig. \ref{viz_ablation} (b)). Object apparent similarity enables the model to accurately perceive and locate all affordance regions in complex backgrounds containing multiple object images (\eg, Fig. \ref{viz_ablation} (c)). Fig. \ref{viz_shp} shows the effect of semantics and pose on the interactive interaction region in the SHP module. This shows that semantics explicitly guides the network to focus on different body parts, while pose guides the network to focus on interaction-related object regions, thereby assisting the network to more accurately mine the interactive affinity.

\par
\begin{figure}[t]
	\centering
		\begin{overpic}[width=0.93\linewidth]{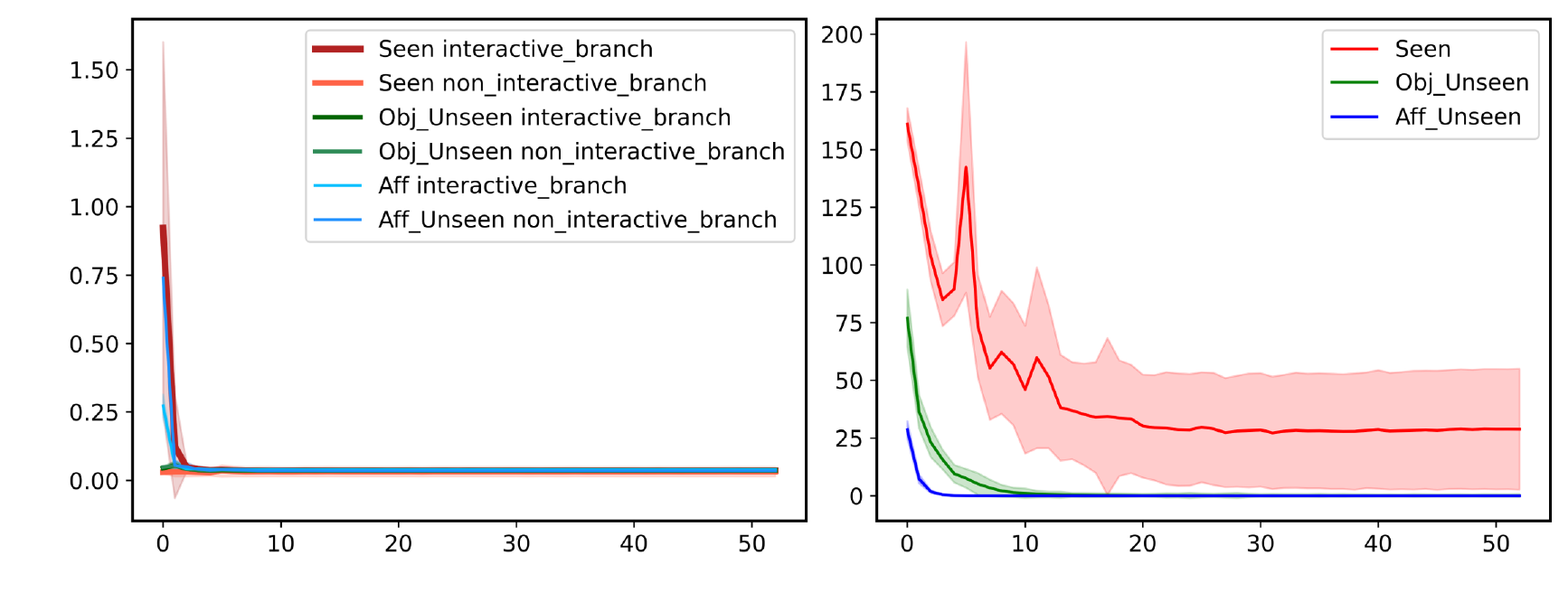}
  \put(0.3,8.6){\rotatebox{90}{{\color{black} \scriptsize{$\bm{||Z-f_\theta(X,Z)||}$}}}}
	\end{overpic}
 \vspace{-13pt}
	\caption{\textbf{Convergence of fixed points.} We show computational convergence results for different DEQ layer indeterminate points.}
	\label{deq2}
\end{figure}

\begin{figure}[t]
	\centering
		\begin{overpic}[width=0.9\linewidth]{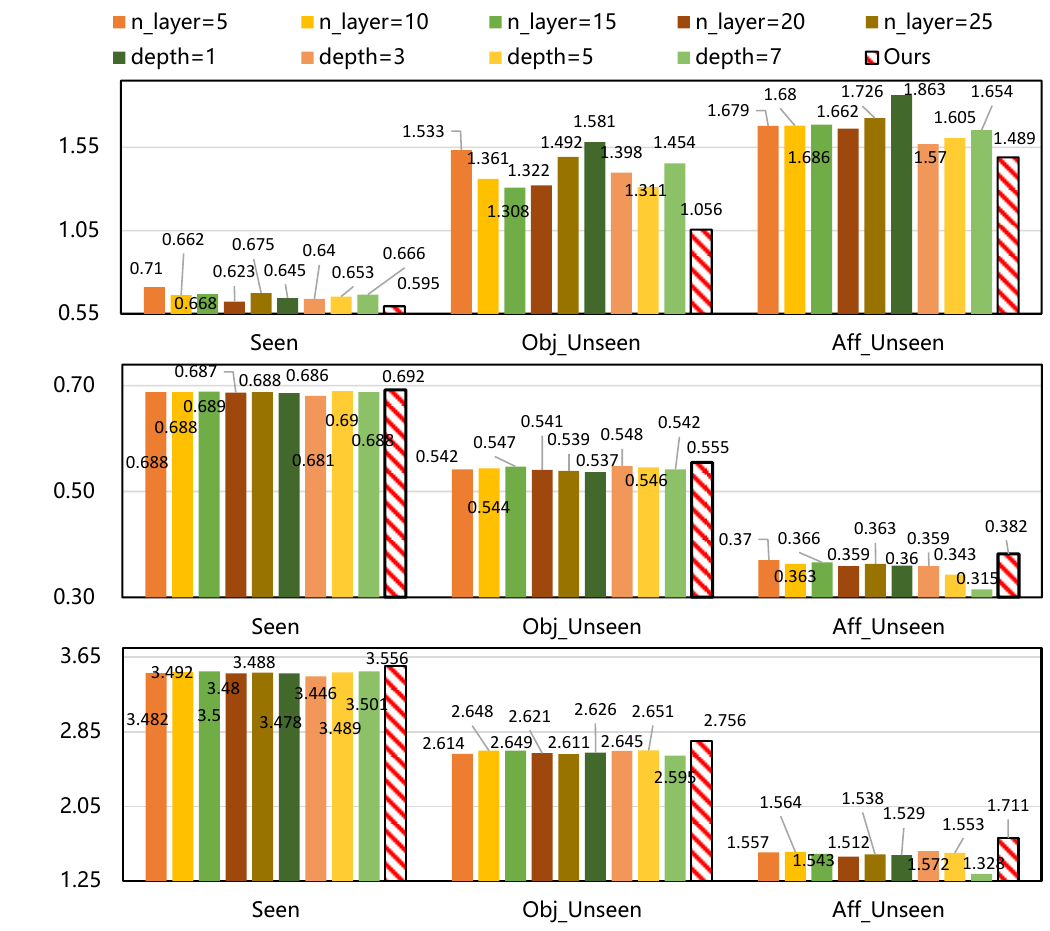}
   \put(0.35,63.7){\rotatebox{90}{{\color{black} \small{\textbf{KLD} $\bm{\downarrow}$}}}}
   \put(0.35,35.6){\rotatebox{90}{{\color{black} \small{\textbf{SIM} $\bm{\uparrow}$}}}}
    \put(0.35,8.6){\rotatebox{90}{{\color{black} \small{\textbf{NSS} $\bm{\uparrow}$}}}}
  
	\end{overpic}
 \vspace{-10pt}
	\caption{\textbf{Different formats of multi-source feature fusion modules.} We investigate other alternatives to the DEQ layer, where $n\_layer=N_1$ means that the input data goes through the same transformer layer $N_1$ times, and $depth=N_2$ represents that the input data passes through the $N_2$ layer transformer for feature fusion. }
	\label{deq3}
\end{figure}

 To explore the dynamic adaptation of the DEQ fusion layer to varying inputs, we visualize the statistics of the number of iterations of the DEQ fuse layer and the results of $||\bm{Z}-f_\theta(\bm{X},\bm{Z})||$, as shown in Fig. \ref{deq1} and Fig. \ref{deq2}. It shows that there is a major discrepancy in the statistical distribution of the number of iterations of the DEQ fusion layer in the all settings, which indicates that it is hard to learn the complex dependency between the pose and the image with a fixed number of layers of the network, whereas the DEQ fusion layer can adaptively adjust the computation according to the inputs of the different branches, which makes the model more stable and mine the corresponding local feature representations. In the Fig. \ref{deq2}, the output of $Z$ converges to a stable state quickly, \ie, the DEQ fusion layer finds a stable and efficient balance, ensuring the stability of the model inference.  Fig. \ref{deq3} shows the results of other multi-source feature fusion approaches. For either a fixed $f_\theta$ or a fixed transformer layer, it is difficult for the model to adaptively adjust the interactions between different pieces of information, leading to pose which makes it difficult to direct the model's focus on the structural cues related to the interactions, thus hindering the model's performance.

\section{Conclusion and Discussion}
\label{sec:Conclusion and Discussion}
This paper proposes to leverage interactive affinity for effective affordance learning by counteracting the influence of multiple possibilities. To this end, a isual-geometric collaborative guided affordance learning  is introduced, which can exploit pose data to guide the network to mine the interactive affinity representation of body parts and object local contact from human-object interaction. Furthermore, we constructed a contact-driven affordance learning (CAL) dataset by collecting and labeling over $55,047$ images from $35$ affordance categories. Our model outperforms eight representative models in three related fields and can serve as a strong baseline for future affordance learning research.

\noindent\textbf{Weakness. } Despite our approach achieving excellent prediction results on affordance learning and a strong generalization, there are still some shortcomings with the model's performance relying more on the similarity of the local representations. This paper introduces pose as a bridge to mine geometrically similar cues, which still perform a relatively smaller role than appearance. We will consider considering cues on geometric structures more explicitly in 3D to achieve more accurate perception.

\noindent\textbf{Future Directions. }  We then discuss several promising future research directions:  \textbf{1) Affordance learning from egocentric videos.} The first-person video can focus more on interaction-related cues \cite{grauman2022ego4d}, understand the intent of human actions \cite{zhang2024pear,zhang2024bidirectional}, and capture interaction details more accurately, enabling finer-grained affordance learning. \textbf{2) Combination of large multimodal models.}  The existing large multimodal models can understand the content in the images and reason accordingly. In the future, we will consider the chain-of-thought method \cite{tang2023cotdet} to further reason about the content in interactive images and extract the interactive affinity representation more accurately. \textbf{3) Collaborative relationship perception.} Human-object interactions are often associated with the collaboration of both hands/feet. The collaborative action between different body parts can more accurately infer the human-object interaction and the object-object interactions in accomplishing a certain task. Thus, future research should incorporate synergies between different human body parts to understand and fully describe human-object interactions.

{\small
\bibliographystyle{IEEEtran}
\bibliography{ref}

\begin{thebibliography}{10}
\providecommand{\url}[1]{#1}
\csname url@samestyle\endcsname
\providecommand{\newblock}{\relax}
\providecommand{\bibinfo}[2]{#2}
\providecommand{\BIBentrySTDinterwordspacing}{\spaceskip=0pt\relax}
\providecommand{\BIBentryALTinterwordstretchfactor}{4}
\providecommand{\BIBentryALTinterwordspacing}{\spaceskip=\fontdimen2\font plus
\BIBentryALTinterwordstretchfactor\fontdimen3\font minus \fontdimen4\font\relax}
\providecommand{\BIBforeignlanguage}[2]{{%
\expandafter\ifx\csname l@#1\endcsname\relax
\typeout{** WARNING: IEEEtran.bst: No hyphenation pattern has been}%
\typeout{** loaded for the language `#1'. Using the pattern for}%
\typeout{** the default language instead.}%
\else
\language=\csname l@#1\endcsname
\fi
#2}}
\providecommand{\BIBdecl}{\relax}
\BIBdecl

\bibitem{luo2023leverage}
H.~Luo, W.~Zhai, J.~Zhang, Y.~Cao, and D.~Tao, ``Leverage interactive affinity for affordance learning,'' in \emph{Proceedings of the IEEE/CVF Conference on Computer Vision and Pattern Recognition}, 2023, pp. 6809--6819.

\bibitem{gibson1977theory}
J.~J. Gibson, ``The theory of affordances,'' \emph{Hilldale, USA}, vol.~1, no.~2, pp. 67--82, 1977.

\bibitem{nagarajan2019grounded}
T.~Nagarajan, C.~Feichtenhofer, and K.~Grauman, ``Grounded human-object interaction hotspots from video,'' in \emph{Proceedings of the IEEE/CVF International Conference on Computer Vision}, 2019, pp. 8688--8697.

\bibitem{nagarajan2020learning}
T.~Nagarajan and K.~Grauman, ``Learning affordance landscapes for interaction exploration in 3d environments,'' \emph{Advances in Neural Information Processing Systems}, vol.~33, pp. 2005--2015, 2020.

\bibitem{qi2017predicting}
S.~Qi, S.~Huang, P.~Wei, and S.-C. Zhu, ``Predicting human activities using stochastic grammar,'' in \emph{Proceedings of the IEEE International Conference on Computer Vision}, 2017, pp. 1164--1172.

\bibitem{chuang2018learning}
C.-Y. Chuang, J.~Li, A.~Torralba, and S.~Fidler, ``Learning to act properly: Predicting and explaining affordances from images,'' in \emph{Proceedings of the IEEE Conference on Computer Vision and Pattern Recognition}, 2018, pp. 975--983.

\bibitem{yamanobe2017brief}
N.~Yamanobe, W.~Wan, I.~G. Ramirez-Alpizar, D.~Petit, T.~Tsuji, S.~Akizuki, M.~Hashimoto, K.~Nagata, and K.~Harada, ``A brief review of affordance in robotic manipulation research,'' \emph{Advanced Robotics}, vol.~31, no. 19-20, pp. 1086--1101, 2017.

\bibitem{chen2015deepdriving}
C.~Chen, A.~Seff, A.~Kornhauser, and J.~Xiao, ``Deepdriving: Learning affordance for direct perception in autonomous driving,'' in \emph{Proceedings of the IEEE international conference on computer vision}, 2015, pp. 2722--2730.

\bibitem{sun2021impression}
Y.~Sun, S.~Fang, and Z.~J. Zhang, ``Impression management strategies on enterprise social media platforms: An affordance perspective,'' \emph{International Journal of Information Management}, vol.~60, p. 102359, 2021.

\bibitem{hassanin2021visual}
M.~Hassanin, S.~Khan, and M.~Tahtali, ``Visual affordance and function understanding: A survey,'' \emph{ACM Computing Surveys (CSUR)}, vol.~54, no.~3, pp. 1--35, 2021.

\bibitem{do2018affordancenet}
T.-T. Do, A.~Nguyen, and I.~Reid, ``Affordancenet: An end-to-end deep learning approach for object affordance detection,'' in \emph{2018 IEEE international conference on robotics and automation (ICRA)}.\hskip 1em plus 0.5em minus 0.4em\relax IEEE, 2018, pp. 5882--5889.

\bibitem{myers2015affordance}
A.~Myers, C.~L. Teo, C.~Ferm{\"u}ller, and Y.~Aloimonos, ``Affordance detection of tool parts from geometric features,'' in \emph{2015 IEEE International Conference on Robotics and Automation (ICRA)}.\hskip 1em plus 0.5em minus 0.4em\relax IEEE, 2015, pp. 1374--1381.

\bibitem{sawatzky2017adaptive}
J.~Sawatzky and J.~Gall, ``Adaptive binarization for weakly supervised affordance segmentation,'' in \emph{Proceedings of the IEEE International Conference on Computer Vision Workshops}, 2017, pp. 1383--1391.

\bibitem{nguyen2016detecting}
A.~Nguyen, D.~Kanoulas, D.~G. Caldwell, and N.~G. Tsagarakis, ``Detecting object affordances with convolutional neural networks,'' in \emph{2016 IEEE/RSJ International Conference on Intelligent Robots and Systems (IROS)}.\hskip 1em plus 0.5em minus 0.4em\relax IEEE, 2016, pp. 2765--2770.

\bibitem{schiavi2022learning}
G.~Schiavi, P.~Wulkop, G.~Rizzi, L.~Ott, R.~Siegwart, and J.~J. Chung, ``Learning agent-aware affordances for closed-loop interaction with articulated objects,'' \emph{arXiv preprint arXiv:2209.05802}, 2022.

\bibitem{ning2024where2explore}
C.~Ning, R.~Wu, H.~Lu, K.~Mo, and H.~Dong, ``Where2explore: Few-shot affordance learning for unseen novel categories of articulated objects,'' \emph{Advances in Neural Information Processing Systems}, vol.~36, 2024.

\bibitem{wang2022adaafford}
Y.~Wang, R.~Wu, K.~Mo, J.~Ke, Q.~Fan, L.~J. Guibas, and H.~Dong, ``Adaafford: Learning to adapt manipulation affordance for 3d articulated objects via few-shot interactions,'' in \emph{European conference on computer vision}.\hskip 1em plus 0.5em minus 0.4em\relax Springer, 2022, pp. 90--107.

\bibitem{luo2021one}
H.~Luo, W.~Zhai, J.~Zhang, Y.~Cao, and D.~Tao, ``One-shot affordance detection,'' \emph{arXiv preprint arXiv:2106.14747}, 2021.

\bibitem{luo2022learning}
------, ``Learning affordance grounding from exocentric images,'' in \emph{Proceedings of the IEEE/CVF conference on computer vision and pattern recognition}, 2022, pp. 2252--2261.

\bibitem{fang2018demo2vec}
K.~Fang, T.-L. Wu, D.~Yang, S.~Savarese, and J.~J. Lim, ``Demo2vec: Reasoning object affordances from online videos,'' in \emph{Proceedings of the IEEE Conference on Computer Vision and Pattern Recognition}, 2018, pp. 2139--2147.

\bibitem{yang2024learning}
F.~Yang, W.~Chen, K.~Yang, H.~Lin, D.~Luo, C.~Tang, Z.~Li, and Y.~Wang, ``Learning granularity-aware affordances from human-object interaction for tool-based functional grasping in dexterous robotics,'' \emph{arXiv preprint arXiv:2407.00614}, 2024.

\bibitem{bai2019deep}
S.~Bai, J.~Z. Kolter, and V.~Koltun, ``Deep equilibrium models,'' \emph{Advances in neural information processing systems}, vol.~32, 2019.

\bibitem{zhai2022one}
W.~Zhai, H.~Luo, J.~Zhang, Y.~Cao, and D.~Tao, ``One-shot object affordance detection in the wild,'' \emph{International Journal of Computer Vision}, vol. 130, no.~10, pp. 2472--2500, 2022.

\bibitem{luo2023grounded}
H.~Luo, W.~Zhai, J.~Zhang, Y.~Cao, and D.~Tao, ``Grounded affordance from exocentric view,'' \emph{International Journal of Computer Vision}, pp. 1--25, 2023.

\bibitem{wang2017binge}
X.~Wang, R.~Girdhar, and A.~Gupta, ``Binge watching: Scaling affordance learning from sitcoms,'' in \emph{Proceedings of the IEEE Conference on Computer Vision and Pattern Recognition}, 2017, pp. 2596--2605.

\bibitem{deng20213d}
S.~Deng, X.~Xu, C.~Wu, K.~Chen, and K.~Jia, ``3d affordancenet: A benchmark for visual object affordance understanding,'' in \emph{Proceedings of the IEEE/CVF Conference on Computer Vision and Pattern Recognition}, 2021, pp. 1778--1787.

\bibitem{sawatzky2017weakly}
J.~Sawatzky, A.~Srikantha, and J.~Gall, ``Weakly supervised affordance detection,'' in \emph{Proceedings of the IEEE Conference on Computer Vision and Pattern Recognition}, 2017, pp. 2795--2804.

\bibitem{liao2022rediscovering}
Y.-C. Liao, K.~Todi, A.~Acharya, A.~Keurulainen, A.~Howes, and A.~Oulasvirta, ``Rediscovering affordance: A reinforcement learning perspective,'' in \emph{CHI Conference on Human Factors in Computing Systems}, 2022, pp. 1--15.

\bibitem{geng2023gapartnet}
H.~Geng, H.~Xu, C.~Zhao, C.~Xu, L.~Yi, S.~Huang, and H.~Wang, ``Gapartnet: Cross-category domain-generalizable object perception and manipulation via generalizable and actionable parts,'' in \emph{Proceedings of the IEEE/CVF Conference on Computer Vision and Pattern Recognition}, 2023, pp. 7081--7091.

\bibitem{geng2023sage}
H.~Geng, S.~Wei, C.~Deng, B.~Shen, H.~Wang, and L.~Guibas, ``Sage: Bridging semantic and actionable parts for generalizable articulated-object manipulation under language instructions,'' \emph{arXiv preprint arXiv:2312.01307}, 2023.

\bibitem{li2024manipllm}
X.~Li, M.~Zhang, Y.~Geng, H.~Geng, Y.~Long, Y.~Shen, R.~Zhang, J.~Liu, and H.~Dong, ``Manipllm: Embodied multimodal large language model for object-centric robotic manipulation,'' in \emph{Proceedings of the IEEE/CVF Conference on Computer Vision and Pattern Recognition}, 2024, pp. 18\,061--18\,070.

\bibitem{Li_2023_CVPR}
G.~Li, V.~Jampani, D.~Sun, and L.~Sevilla-Lara, ``Locate: Localize and transfer object parts for weakly supervised affordance grounding,'' in \emph{Proceedings of the IEEE/CVF Conference on Computer Vision and Pattern Recognition (CVPR)}, June 2023, pp. 10\,922--10\,931.

\bibitem{Yang_2023_ICCV}
Y.~Yang, W.~Zhai, H.~Luo, Y.~Cao, J.~Luo, and Z.-J. Zha, ``Grounding 3d object affordance from 2d interactions in images,'' in \emph{Proceedings of the IEEE/CVF International Conference on Computer Vision (ICCV)}, October 2023, pp. 10\,905--10\,915.

\bibitem{Chen_2023_CVPR}
J.~Chen, D.~Gao, K.~Q. Lin, and M.~Z. Shou, ``Affordance grounding from demonstration video to target image,'' in \emph{Proceedings of the IEEE/CVF Conference on Computer Vision and Pattern Recognition (CVPR)}, June 2023, pp. 6799--6808.

\bibitem{Gkioxari_2018_CVPR}
G.~Gkioxari, R.~Girshick, P.~Dollár, and K.~He, ``Detecting and recognizing human-object interactions,'' in \emph{Proceedings of the IEEE Conference on Computer Vision and Pattern Recognition (CVPR)}, June 2018.

\bibitem{chao2018learning}
Y.-W. Chao, Y.~Liu, X.~Liu, H.~Zeng, and J.~Deng, ``Learning to detect human-object interactions,'' in \emph{2018 IEEE Winter Conference on Applications of Computer Vision (WACV)}.\hskip 1em plus 0.5em minus 0.4em\relax IEEE, 2018, pp. 381--389.

\bibitem{Kim_2021_CVPR}
B.~Kim, J.~Lee, J.~Kang, E.-S. Kim, and H.~J. Kim, ``Hotr: End-to-end human-object interaction detection with transformers,'' in \emph{Proceedings of the IEEE/CVF Conference on Computer Vision and Pattern Recognition (CVPR)}, June 2021, pp. 74--83.

\bibitem{shimada2022hulc}
S.~Shimada, V.~Golyanik, Z.~Li, P.~P{\'e}rez, W.~Xu, and C.~Theobalt, ``Hulc: 3d human motion capture with pose manifold sampling and dense contact guidance,'' in \emph{European Conference on Computer Vision}.\hskip 1em plus 0.5em minus 0.4em\relax Springer, 2022, pp. 516--533.

\bibitem{bhatnagar2022behave}
B.~L. Bhatnagar, X.~Xie, I.~A. Petrov, C.~Sminchisescu, C.~Theobalt, and G.~Pons-Moll, ``Behave: Dataset and method for tracking human object interactions,'' in \emph{Proceedings of the IEEE/CVF Conference on Computer Vision and Pattern Recognition}, 2022, pp. 15\,935--15\,946.

\bibitem{mao2022contact}
W.~Mao, M.~Liu, R.~Hartley, and M.~Salzmann, ``Contact-aware human motion forecasting,'' \emph{arXiv preprint arXiv:2210.03954}, 2022.

\bibitem{yang2021cpf}
L.~Yang, X.~Zhan, K.~Li, W.~Xu, J.~Li, and C.~Lu, ``Cpf: Learning a contact potential field to model the hand-object interaction,'' in \emph{Proceedings of the IEEE/CVF International Conference on Computer Vision}, 2021, pp. 11\,097--11\,106.

\bibitem{xie2021segformer}
E.~Xie, W.~Wang, Z.~Yu, A.~Anandkumar, J.~M. Alvarez, and P.~Luo, ``Segformer: Simple and efficient design for semantic segmentation with transformers,'' \emph{Advances in Neural Information Processing Systems}, vol.~34, pp. 12\,077--12\,090, 2021.

\bibitem{dosovitskiy2020image}
A.~Dosovitskiy, L.~Beyer, A.~Kolesnikov, D.~Weissenborn, X.~Zhai, T.~Unterthiner, M.~Dehghani, M.~Minderer, G.~Heigold, S.~Gelly \emph{et~al.}, ``An image is worth 16x16 words: Transformers for image recognition at scale,'' in \emph{International Conference on Learning Representations}, 2020.

\bibitem{turc2019}
I.~Turc, M.-W. Chang, K.~Lee, and K.~Toutanova, ``Well-read students learn better: On the importance of pre-training compact models,'' \emph{arXiv preprint arXiv:1908.08962v2}, 2019.

\bibitem{cai2024smpler}
Z.~Cai, W.~Yin, A.~Zeng, C.~Wei, Q.~Sun, W.~Yanjun, H.~E. Pang, H.~Mei, M.~Zhang, L.~Zhang \emph{et~al.}, ``Smpler-x: Scaling up expressive human pose and shape estimation,'' \emph{Advances in Neural Information Processing Systems}, vol.~36, 2024.

\bibitem{vaswani2017attention}
A.~Vaswani, N.~Shazeer, N.~Parmar, J.~Uszkoreit, L.~Jones, A.~N. Gomez, {\L}.~Kaiser, and I.~Polosukhin, ``Attention is all you need,'' \emph{Advances in neural information processing systems}, vol.~30, 2017.

\bibitem{lin2014microsoft}
T.-Y. Lin, M.~Maire, S.~Belongie, J.~Hays, P.~Perona, D.~Ramanan, P.~Doll{\'a}r, and C.~L. Zitnick, ``Microsoft coco: Common objects in context,'' in \emph{Proceedings of the European Conference on Computer Vision (ECCV)}.\hskip 1em plus 0.5em minus 0.4em\relax Springer, 2014, pp. 740--755.

\bibitem{krishna2017visual}
R.~Krishna, Y.~Zhu, O.~Groth, J.~Johnson, K.~Hata, J.~Kravitz, S.~Chen, Y.~Kalantidis, L.-J. Li, D.~A. Shamma \emph{et~al.}, ``Visual genome: Connecting language and vision using crowdsourced dense image annotations,'' \emph{International journal of computer vision}, vol. 123, pp. 32--73, 2017.

\bibitem{bylinskii2015saliency}
Z.~Bylinskii, T.~Judd, A.~Borji, L.~Itti, F.~Durand, A.~Oliva, and A.~Torralba, ``Mit saliency benchmark,'' 2015.

\bibitem{judd2012benchmark}
T.~Judd, F.~Durand, and A.~Torralba, ``A benchmark of computational models of saliency to predict human fixations,'' 2012.

\bibitem{gebru2021datasheets}
T.~Gebru, J.~Morgenstern, B.~Vecchione, J.~W. Vaughan, H.~Wallach, H.~D. Iii, and K.~Crawford, ``Datasheets for datasets,'' \emph{Communications of the ACM}, vol.~64, no.~12, pp. 86--92, 2021.

\bibitem{zhao2017pyramid}
H.~Zhao, J.~Shi, X.~Qi, X.~Wang, and J.~Jia, ``Pyramid scene parsing network,'' in \emph{Proceedings of the IEEE conference on computer vision and pattern recognition}, 2017, pp. 2881--2890.

\bibitem{chen2018encoder}
L.-C. Chen, Y.~Zhu, G.~Papandreou, F.~Schroff, and H.~Adam, ``Encoder-decoder with atrous separable convolution for semantic image segmentation,'' in \emph{Proceedings of the European conference on computer vision (ECCV)}, 2018, pp. 801--818.

\bibitem{xu2022vitpose}
Y.~Xu, J.~Zhang, Q.~Zhang, and D.~Tao, ``Vi{TP}ose: Simple vision transformer baselines for human pose estimation,'' in \emph{Advances in Neural Information Processing Systems}, 2022.

\bibitem{YuanFHLZCW21}
Y.~Yuan, R.~Fu, L.~Huang, W.~Lin, C.~Zhang, X.~Chen, and J.~Wang, ``Hrformer: High-resolution transformer for dense prediction,'' 2021.

\bibitem{min2021hypercorrelation}
J.~Min, D.~Kang, and M.~Cho, ``Hypercorrelation squeeze for few-shot segmentation,'' in \emph{Proceedings of the IEEE/CVF International Conference on Computer Vision (ICCV)}, 2021.

\bibitem{radford2021learning}
A.~Radford, J.~W. Kim, C.~Hallacy, A.~Ramesh, G.~Goh, S.~Agarwal, G.~Sastry, A.~Askell, P.~Mishkin, J.~Clark \emph{et~al.}, ``Learning transferable visual models from natural language supervision,'' in \emph{International conference on machine learning}.\hskip 1em plus 0.5em minus 0.4em\relax PMLR, 2021, pp. 8748--8763.

\bibitem{bylinskii2018different}
Z.~Bylinskii, T.~Judd, A.~Oliva, A.~Torralba, and F.~Durand, ``What do different evaluation metrics tell us about saliency models?'' \emph{IEEE transactions on pattern analysis and machine intelligence}, vol.~41, no.~3, pp. 740--757, 2018.

\bibitem{swain1991color}
M.~J. Swain and D.~H. Ballard, ``Color indexing,'' \emph{International Journal of Computer Vision (IJCV)}, vol.~7, no.~1, pp. 11--32, 1991.

\bibitem{peters2005components}
R.~J. Peters, A.~Iyer, L.~Itti, and C.~Koch, ``Components of bottom-up gaze allocation in natural images,'' \emph{Vision research}, vol.~45, no.~18, pp. 2397--2416, 2005.

\bibitem{loshchilov2017decoupled}
I.~Loshchilov and F.~Hutter, ``Decoupled weight decay regularization,'' \emph{arXiv preprint arXiv:1711.05101}, 2017.

\bibitem{grauman2022ego4d}
K.~Grauman, A.~Westbury, E.~Byrne, Z.~Chavis, A.~Furnari, R.~Girdhar, J.~Hamburger, H.~Jiang, M.~Liu, X.~Liu \emph{et~al.}, ``Ego4d: Around the world in 3,000 hours of egocentric video,'' in \emph{Proceedings of the IEEE/CVF Conference on Computer Vision and Pattern Recognition}, 2022, pp. 18\,995--19\,012.

\bibitem{zhang2024pear}
Z.~Zhang, H.~Luo, W.~Zhai, Y.~Cao, and Y.~Kang, ``Pear: Phrase-based hand-object interaction anticipation,'' \emph{arXiv preprint arXiv:2407.21510}, 2024.

\bibitem{zhang2024bidirectional}
------, ``Bidirectional progressive transformer for interaction intention anticipation,'' \emph{arXiv preprint arXiv:2405.05552}, 2024.

\bibitem{tang2023cotdet}
J.~Tang, G.~Zheng, J.~Yu, and S.~Yang, ``Cotdet: Affordance knowledge prompting for task driven object detection,'' in \emph{Proceedings of the IEEE/CVF International Conference on Computer Vision}, 2023, pp. 3068--3078.

\end{thebibliography}
}

\vfill

%\appendix

\end{document}